\documentclass[aoas,preprint]{imsart}

\RequirePackage{amsthm,amsmath,amsfonts,amssymb}
\RequirePackage[authoryear]{natbib}

\usepackage{xcolor}
\definecolor{darkblue}{rgb}{0.0, 0.0, 0.55}
\RequirePackage[colorlinks,citecolor=darkblue,urlcolor=blue]{hyperref}

\usepackage[utf8]{inputenc} 
\usepackage[T1]{fontenc}    
\usepackage{url}            
\usepackage{booktabs}       
\usepackage{amsfonts}       
\usepackage{nicefrac}       
\usepackage{microtype}      
\usepackage{lipsum}
\usepackage{amsmath}

\usepackage[ruled,vlined]{algorithm2e}
\usepackage{algorithmic}

\usepackage{graphicx}
\usepackage{caption, subcaption}

\usepackage{amsthm}

\DeclareMathOperator*{\argmin}{arg\,min}

\theoremstyle{definition}
\newtheorem{definition}{Definition}[section]

\usepackage{soul}

\usepackage[inline]{enumitem}

\newcommand{\Romannumeral}[1]{\MakeUppercase{\romannumeral #1}}

\newcommand{\beginsupplement}{%
        \setcounter{section}{0}
        \renewcommand{\thesection}{S\arabic{section}}%
        \setcounter{table}{0}
        \renewcommand{\thetable}{S\arabic{table}}%
        \setcounter{figure}{0}
        \renewcommand{\thefigure}{S\arabic{figure}}%
     }

\startlocaldefs

\endlocaldefs

\begin{document}

\begin{frontmatter}
\title{Causal Network Learning with Non-invertible \\ Functional Relationships}

\begin{aug}
\author{\fnms{Bingling} \snm{Wang}\thanks{UCLA Department of Biostatistics, binglingwang@ucla.edu}}
\and
\author{\fnms{Qing} \snm{Zhou}\thanks{UCLA Department of Statistics, zhou@stat.ucla.edu}}


\end{aug}

\begin{abstract}
Discovery of causal relationships from observational data is an important problem in many areas. Several recent results have established the identifiability of causal DAGs with non-Gaussian and/or nonlinear structural equation models (SEMs). 
In this paper, we focus on nonlinear SEMs defined by non-invertible functions, which exist in many data domains, and propose a novel test for non-invertible bivariate causal models. We further develop a method to incorporate this test in structure learning of DAGs that contain both linear and nonlinear causal relations. By extensive numerical comparisons, we show that our algorithms outperform existing DAG learning methods in identifying causal graphical structures. We illustrate the practical application of our method in learning causal networks for combinatorial binding of transcription factors from ChIP-Seq data.
\end{abstract}


\end{frontmatter}


\section{Introduction}

Inferring causal relations from data is a fundamental problem in many areas of science. 
Randomized controlled experiments are the gold standard tool used for causal discovery. However, there are certain limitations, such as expenses, time, ethics, practicalities etc, in the application of randomized experiments. Even when experiments are possible to carry out, with hundreds and thousands of variables easily collected nowadays, performing a large number of experiments on these variables is unrealistic when background knowledge is limited. 
Identifying causal relationships from observational data has therefore attracted much attention from many researchers in the past few decades [\cite{Verma1990,  Meek1995, Chickering1996, HeckermanBook2006, pearl2009, Spirtes2010}]. 

In this paper, we model causal relations among a set of random variables by a directed acyclic graph (DAG), following \cite{pearl2009}. Under this approach, causal structure learning is achieved by estimating the structure of the underlying causal DAG from observed data. Traditional structure learning methods can be classified into two categories. The first category is the constraint-based approach which seeks to recover the underlying graphical structure by identifying conditional independence relationships between variables. Examples of constraint-based algorithms include the PC algorithm by \cite{Spirtes1991} and the Fast Causal Inference (FCI) algorithm by \cite{Spirtes2000}. The second category is the score-based approach that aims to find the causal DAG by maximizing certain scoring function, e.g. Bayesian Dirichlet scores, Bayesian information criterion, or regularized likelihood among others. Algorithms in this category, such as the Greedy Equivalence Search (GES) by \cite{Chickering2003} and coordinate descent by \cite{Fu2013}, search the space of graphs for an optimal structure using greedy, local, or some other search strategies.

For continuous data, most existing DAG learning methods assume linear parent-child relations with additive Gaussian noises, sometimes called linear Gaussian DAGs [\cite{pearl2009, Spirtes2000}]. Although models under assumptions of linearity and Gaussianity are well understood and convenient to work with, they are not always realistic in real-world applications. It is arguable that most causal relations in real data are more or less nonlinear in nature. Moreover, linear Gaussian DAGs are not identifiable from observational data. Every DAG in the Markov equivalence class of the true causal DAG gives identical likelihood of observational data and implies an identical set of conditional independence relations. Therefore, neither constraint-based nor score-based approaches can identify the causal DAG. This is the well-known non-identifiability issue of linear Gaussian DAGs. Consider a simple problem of inferring the causality between two variables, whether $X$ causes $Y$ or vice versa. In terms of DAGs, we are considering either $X \to Y$ or $Y \to X$. Under linear and Gaussian assumptions, the two DAGs merely represents two ways to factorize the same bivariate Gaussian density $p(y|x)p(x) = p(x|y)p(x)$, and thus one cannot distinguish the two causal models from observational data in this case. 

In recent years, many efforts have been made to tackle the causal discovery problem from different perspectives under various identifiability assumptions [\cite{Shimizu2006, Zhang2008, Zhang2009, Hoyer2009, Mooij2010, Shimizu2011, Peters2014, Peters2014b, Peters2016, Blobaum2018, Monti2019}]. In particular, a few methods have been proposed to identify the true causal DAG from observational data by making use of nonlinear and/or non-Gaussian structural equation models (SEMs). \cite{Shimizu2006} showed that the true causal DAG is identifiable assuming non-Gaussian errors under linear SEMs and proposed a linear non-Gaussian acyclic model (LiNGAM) for causal structure learning. \cite{Hoyer2009} pointed out that nonlinearity can break the symmetry between observed variables, which leads to identifiable causal models, and proposed a nonlinear additive noise model which was further extended and implemented by \cite{Peters2014}. \cite{Zhang2009} proposed a post-nonlinear (PNL) causal model under which one can distinguish the cause from effect and investigated conditions for identifiability of the model. These recent developments are reviewed in \cite{Mooij2016} and \cite{Glymour2019}. 

A key ingredient in the above methods is the use of general independence tests to determine the causal directions [\cite{Shimizu2006,Shimizu2011,Peters2014,Mooij2016}]. Take the simple bivariate case as an example. Suppose the true causal model is $X\to Y$ so that the corresponding SEM is $Y=f(X)+\epsilon$, where the noise $\epsilon$ is independent of the causal parent $X$. If $f$ is a nonlinear function satisfying some mild conditions, one cannot find a function $g(\cdot)$ such that $X=g(Y)+\epsilon'$ and that $\epsilon'$ is independent of $Y$. Thus, to identify the correct causal DAG, one must test whether the residual after a nonlinear regression of $Y$ onto $X$ is independent of $X$. This is in general a very difficult problem, since in many regression techniques the residual is uncorrelated with $X$ by design. Thus, advanced and complex test procedures such as the Hilbert-Schmidt Independence Criterion (HSIC) [\cite{gretton2005kernel}], a kernel-based independence test, is often used in this approach. To estimate a causal DAG on many variables, a sequence of such independence tests is usually performed by these methods to identify a causal ordering and the parent set of each variable. 

In this paper, we restrict our attention to non-invertible causal relations between variables in a DAG, which has not been explored in the literature. In the bivariate case, we assume the function $f$ is not invertible. We develop a novel method to identify the causal direction by test for non-invertibility of $f$. This is less general than the above methods that apply to many nonlinear functions, however, our approach can be more powerful for the problem we consider and does not rely on complicated independence tests. Moreover, we assume that the causal relations in a DAG are a mix of linear and nonlinear relationships. Accordingly, we propose a few approaches to combine linear structure learning methods, such as the PC algorithm, with our non-invertible function identification in a principled way to estimate the full causal DAG structure. Our numerical comparisons show that our combined approach outperforms both the traditional linear structure learning methods and the recent nonlinear DAG learning methods.

Our causal learning method is widely applicable to many data domains. First, causal structure estimation under the DAG framework has become popular in different applied fields, including genomics [\cite{Sachs2005, Gao2015}], epidemiology [\cite{Greenland1999, Joffe2012}] and social sciences [\cite{Velikova2014, Garvey2015}]. Second, identification of nonlinear and non-invertible causal relations by our method will bring new insights into the underlying scientific problem. Most graphical model approaches to large-scale problems work under linearity assumptions, which serve as a good approximation if the underlying nonlinear relationship is monotone and close to a linear function. These methods are usually not sensitive enough to identify non-invertible causal relations. Our method fills this gap. A non-invertible relationship can review complicated causality among the variables of interest. Use gene regulation as an example. The expression of a gene is often regulated by the binding of multiple proteins, called transcription factors (TFs), to the upstream DAG sequence of the gene. The presence or absence of one TF $X$ may cause a change of the binding of another TF $Y$. Such causality among the binding activities among a set of TFs may be reviewed by learning a DAG from their binding data, in particular, ChIP-Seq data. 
There could be nonlinear relations in this problem, which reflects the complexity in combinatorial gene regulation. We will apply our method to ChIP-Seq data to demonstrate its use in scientific discovery.

The remainder of this paper is organized as follows. We start with introducing our bivariate non-invertible SEM and test of causal direction in Section~\ref{sec:causal_bivariate}. We then incorporate this method into structure learning of causal networks with both linear and nonlinear SEMs in Section~\ref{sec:causal_structure_learning}. Section~\ref{sec:experiments} evaluates the performance of the proposed algorithms under different simulation settings and compares with other competing DAG learning methods. Section~\ref{sec:application} presents an application to ChIP-Seq data for the construction of a TF binding causal network. The paper concludes with discussions in Section~\ref{sec:discussion}. In supplementary material, we provide some technical details of our algorithms and additional numerical results. 

\section{Non-invertible bivariate causal relations} \label{sec:causal_bivariate}

\subsection{Bivariate non-invertible SEM}\label{ss:bivariate_NISEM} 

Consider two random variables $X$ and $Y$ that may be causally related. Our task here is to decide whether there is indeed a causal relation between the two variables and if so whether the relation is $X \to Y$ (i.e. $X$ causes $Y$) or $Y \to X$. We assume only observational data are available. To make the causal relation identifiable from observational data, we will consider a non-invertible SEM between $X$ and $Y$ defined as follows.

\begin{definition} \label{def:nisem}
Suppose two random variables $X$ and $Y$ satisfy a nonlinear SEM, $Y = f(X) + \epsilon$, where $\epsilon$ is independent of $X$ and the function $f$ is non-trivial (i.e. not a constant function) and non-invertible (i.e. $f^{-1}(\cdot)$ does not exist). Then we say that $X$ and $Y$ follow a bivariate non-invertible SEM (NISEM), which defines the causal relation $X \to Y$.
\end{definition}
It follows immediately from the identifiability of nonlinear SEMs [\cite{Zhang2009}, Corollary 10] that a bivariate NISEM is identifiable.

For now, we assume that either $X \to Y$ or $Y \to X$, or they are not causally related so that $X$ is independent of $Y$. To infer the causal relation between $X$ and $Y$, 
we consider three scenarios accordingly: 
\begin{enumerate}
    \item $X$ and $Y$ are independent. We infer that there is no causal relationship between $X$ and $Y$.
    \item $X$ and $Y$ are dependent, but the function $f$ is invertible. In this case there exists causality between $X$ and $Y$, but we are not able to identify the direction of the relation.
    \item $X$ and $Y$ are dependent, and the function $f$ is non-invertible. We conclude that there exists causal relationship between $X$ and $Y$ and we are able to determine the causal direction.
\end{enumerate}

Given observational data, we develop a method to determine which of  the above three cases is supported by the data. Our main method is a two-step approach: First, test whether $X$ and $Y$ are statistically independent. If the two variables are not independent, we then continue to fit a bivariate nonlinear SEM for $(X,Y)$ and test whether the function $f$ is invertible. Although the general identifiability results in \cite{Peters2014} apply to a large class of nonlinear functions, including invertible functions, our primary focus in this work is on the non-invertible cases. By limiting to non-invertible functions, our method gains substantial increase in power and accuracy, as demonstrated in our numerical comparisons in Section~\ref{ss:ANM_comparison} 

Many non-invertible functions can be well approximated by piecewise linear functions. Here, we use a piecewise linear function with two pieces to approximate the functional relation between $X$ and $Y$. That is, we assume 
\begin{equation*}
{f}(x) =
\begin{cases}
a_l + b_l x & x \leq {\tau}_x \\
a_h + b_h x & x > {\tau}_x \\
\end{cases},
\end{equation*}
where $\tau_x$ is the cut point between the two pieces of linear functions and $a_l, b_l, a_h, b_h$ are coefficients of the linear functions.

\begin{figure}
    \centering
    \includegraphics[width= 0.5\linewidth]{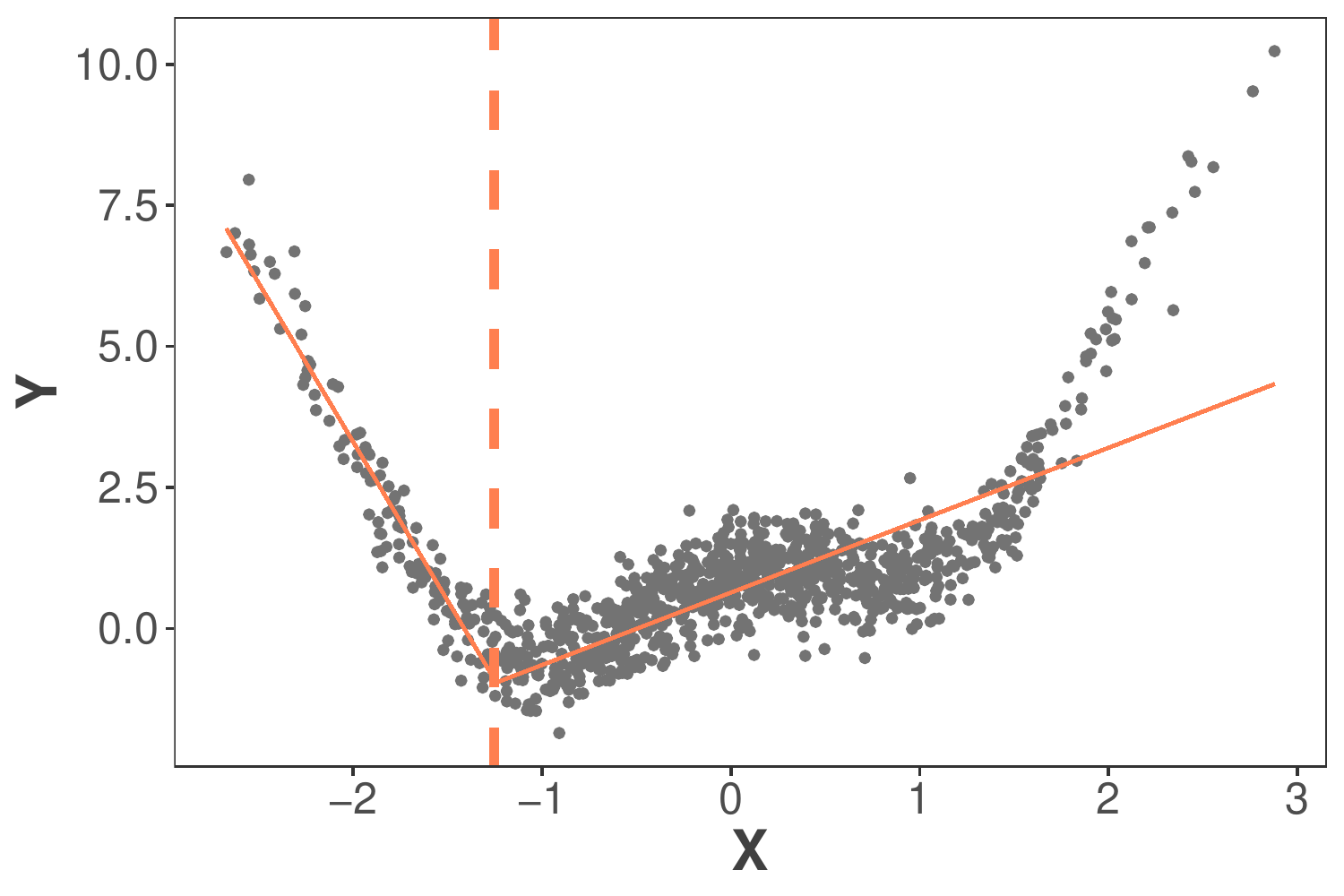}
    \caption{Approximating a nonlinear relationship by piecewise linear function with two pieces. The cut point is indicated by the vertical dashed line.}  \label{fig:two_piece_demo}  
\end{figure}

Our motivation to use a piecewise linear function stems from the fact that nonlinear relationships can generally be approximated using a sufficient number of pieces. A non-invertible function by definition is not monotonic and a piecewise linear function can easily capture the nonlinear trend with a well-chosen cutoff point. For example, a quadratic function can be approximated by two pieces of linear functions, each having a very different slope. For nonlinear relationships with multiple tuning points, it is more accurate to use multiple pieces of linear functions. However, determining the number of pieces and fitting a many-piece model can be inaccurate and may increase the risk of overfitting in practice. Fortunately, simply capturing two linear pieces with a significant change in their slopes is sufficient for our purpose of detecting the causal direction in a non-invertible relationship. See Figure~\ref{fig:two_piece_demo} for an illustration. 
Simulation results on different nonlinear patterns in Section~\ref{ss:ANM_comparison} confirm the robustness of our approach. On the other hand, it is possible to generalize our method to allow multiple linear pieces with a change point detection procedure [e.g. \cite{Pettitt1979, Reeves2007, Chen2014}], which will be left for future work.

\subsection{Model fitting}\label{ss:model_fitting}

We will first discuss how to fit a piecewise linear function from data and then propose a statistic to measure the goodness of fit, which will be used in our determination of the causal direction in next subsection.

Suppose we have observed data $(\mathbf{x},\mathbf{y})=\{(x_i,y_i):i=1,\ldots,n\}$, an i.i.d. sample from the joint distribution of $(X,Y)$. Assuming the causal direction is $X\to Y$, a corresponding bivariate nonlinear SEM is estimated in the following way. First we find the cut point $\tau_x$ of the piecewise function. We restrict the domain of $\tau_x$ to be a set of quantiles of $\mathbf{x}$, denoted by $T = \{t_j, j = 1, ..., m\}$. For each $t_j\in T$, fit a piecewise linear function which yields two residual sums of squares $\mathit{rss_l(t_j)}$ for $x \leq t_j$, and $\mathit{rss_h}(t_j)$ for $x > t_j$. Then the estimate of $\tau_x$ is found by minimizing the total residual sum of squares of the two segments:
\begin{equation} \label{eq:tau}
 \hat{\tau}_x = \argmin_{t\in T} \{\mathit{rss_l}(t) + \mathit{rss_h}(t)\}.   
\end{equation}
Second, given $\hat{\tau}_x$ we fit a linear function in each segment. Write the estimated $\hat{f}$ as
\begin{equation} \label{eq:piecewise_function}
\hat{f}(x) =
\begin{cases}
\hat{a}_l + \hat{b}_l x & x \leq \hat{\tau}_x \\
\hat{a}_h + \hat{b}_h x & x > \hat{\tau}_x \\
\end{cases}, 
\end{equation}
where $\hat{\tau}_x$ is the estimated cut point, and $\hat{a}_l, \hat{b}_l$, $\hat{a}_h, \hat{b}_h$ are the estimated coefficients for the two linear segments.

To measure the goodness of fit of the piecewise model, we define a statistic $\bar R^2$, which is a weighted average of the $R^2$ of each piece:
\begin{align}\label{eq:R2}
    \bar R^2_{X \to Y} = \frac{n_l r_l^2 + n_h^2 r_h^2}{n_l+n_h},
\end{align}
where $n_l$ and $n_h$ are the number of observations in the subsets $\{ i: x_i \leq \hat{\tau}_x \}$ and $\{ i: x_i > \hat{\tau}_x \}$, respectively; $r_l, r_h$ are the corresponding sample correlation coefficients between $x_i$ and $y_i$.

To test whether $f$ is non-invertible, we will swap $x$ and $y$ in the above procedure to fit a nonlinear SEM for $Y \to X$ and evaluate the model fitting by calculating $\bar R^2_{Y \to X}$. Then we will design a test to decide the causal direction between $X$ and $Y$ based on $\bar R^2_{X \to Y}$ and $\bar R^2_{Y \to X}$.

\subsection{Test for causal direction}\label{ss:hypothesis_testing}

Although one direction may be preferred than the other based on the model fitting statistics, we need to find out whether this preference is statistically significant. Thus, hypothesis testing is necessary to decide whether the function $f$ is indeed non-invertible. Our null hypothesis $H_0$ is that the functional relationship $f$ between $X$ and $Y$ is invertible.

We define a test statistic
\begin{align}\label{eq:test_stat}
\eta = \max\big\{\bar{R}^2_{X \to Y} / \bar{R}^2_{Y \to X}, \bar{R}^2_{Y \to X} / \bar{R}^2_{X \to Y}\big\}
\end{align}
to compare the goodness of fit between the two nonlinear SEMs.
Under $H_0$ that $f$ is invertible, the two SEMs would fit the data equally well so that the values of $\bar{R}^2$ for the two nonlinear SEMs will be close to each other. Therefore, the corresponding $\eta$ should be close to $1$.  Under $H_a$ that $f$ is non-invertible, the values of $\bar{R}^2$ for the two nonlinear SEMs will be significantly different and $\eta$ will be significantly greater than 1. Let $\eta_0$ be a random variable following the distribution of $\eta$ under $H_0$, and $\hat{\eta}$ be the observed value of the comparison statistic $\eta$. The $p$-value of the hypothesis test is $P(\eta_0 \geq \hat{\eta} | H_0)$. Now the question is how to obtain the distribution of $\eta_0$. We propose two different methods to approximate this null distribution.

\begin{figure}
    \centering
    \begin{subfigure}[t]{0.48\linewidth}
    \centering\includegraphics[width=\linewidth]{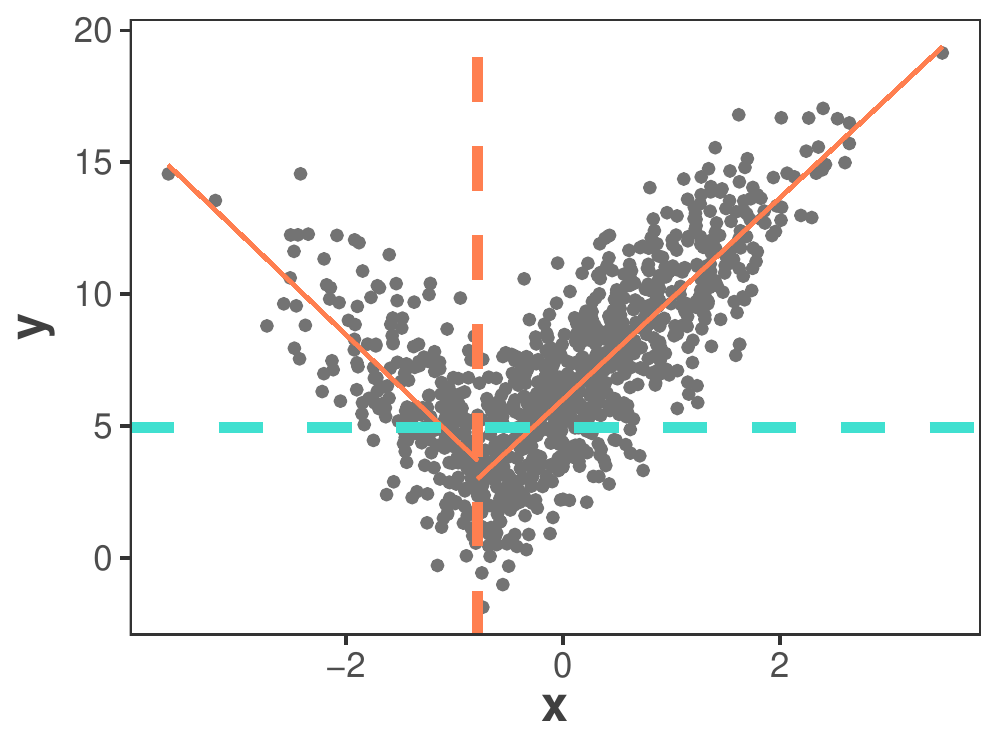} 
    \caption{Observed data and the fitted piecewise linear functions \label{fig:hypothesis_test_a}}
    \end{subfigure}
    \begin{subfigure}[t]{0.48\linewidth}
    \centering\includegraphics[width=\linewidth]{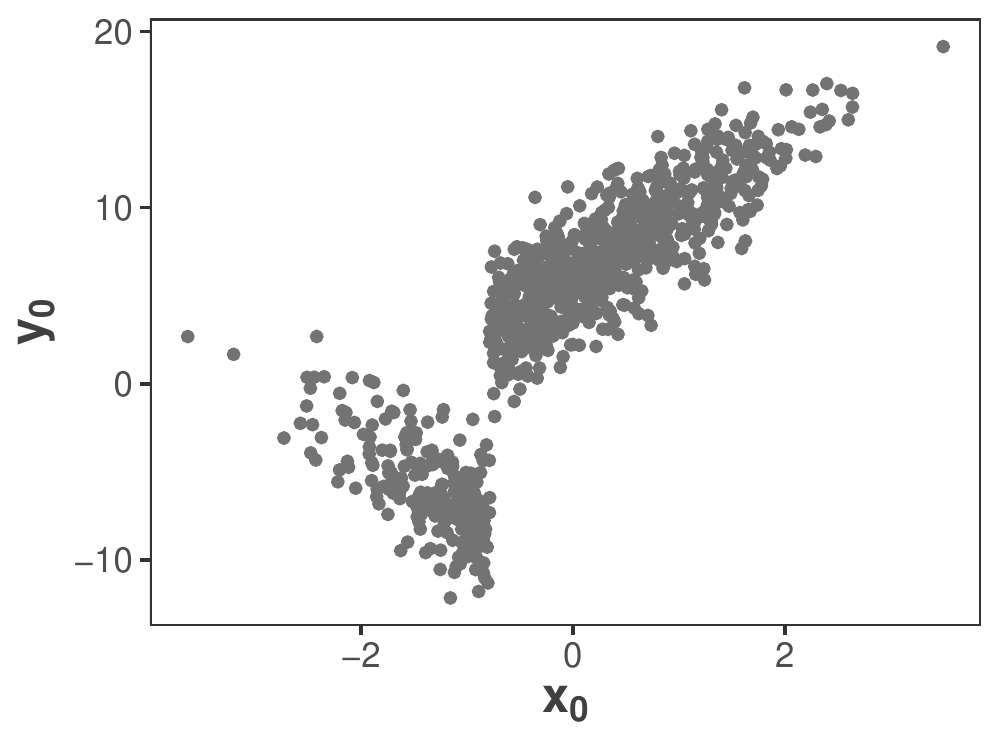}
    \caption{Modified data under null hypothesis \label{fig:hypothesis_test_b}}
    \end{subfigure}
    
    \vspace{2ex}
    \begin{subfigure}[t]{0.48\linewidth}
    \centering\includegraphics[width=\linewidth]{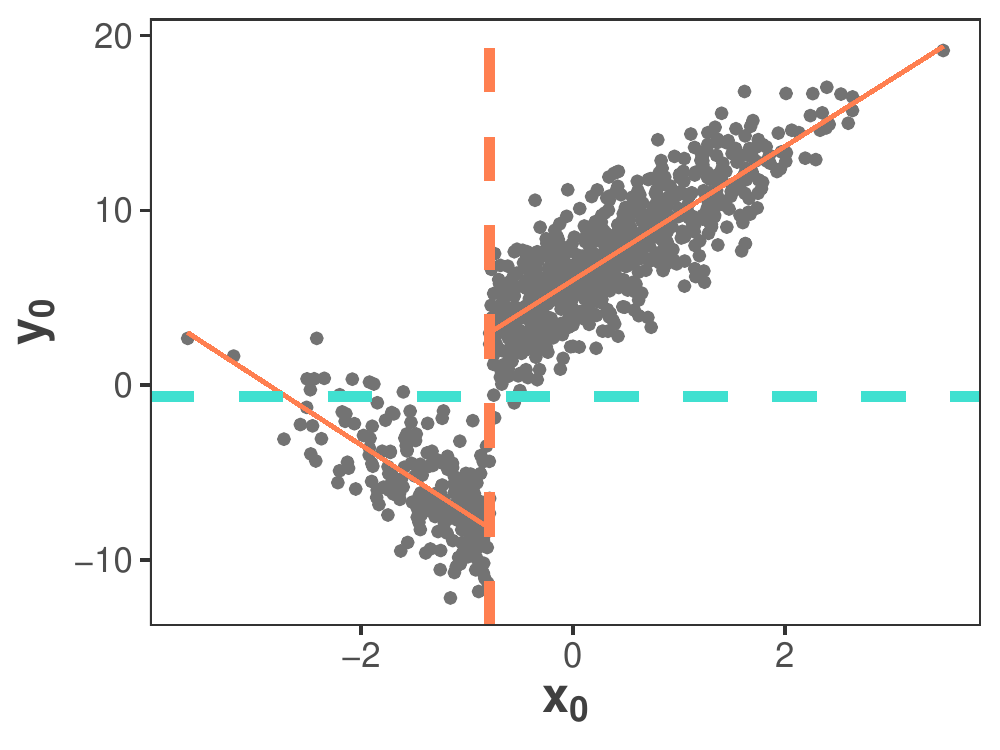}
    \caption{Fitted piecewise linear functions for $x_0 \to y_0$ \label{fig:hypothesis_test_c}}
    \end{subfigure}
    \begin{subfigure}[t]{0.48\linewidth}
    \centering\includegraphics[width=\linewidth]{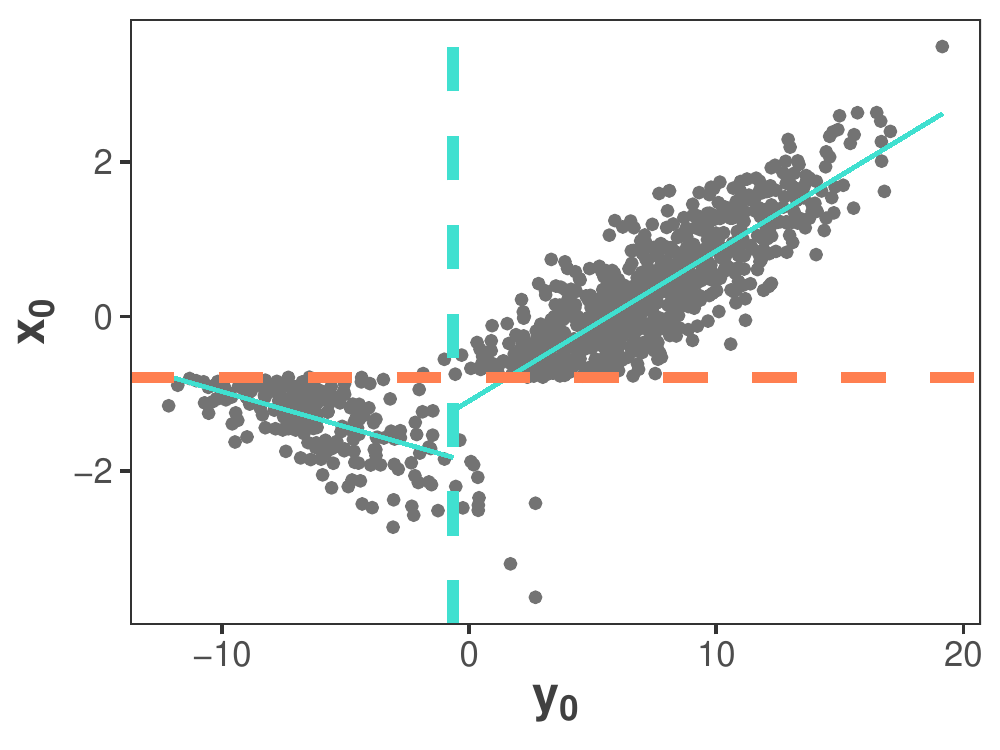}
    \caption{Fitted piecewise linear functions for $y_0 \to x_0$ \label{fig:hypothesis_test_d}}
    \end{subfigure}

    \caption{Illustration of test for causal direction. Solid lines are the fitted piecewise linear functions; dashed lines indicate the cut points found in the two variables.}
    \label{fig:hypothesis_test}
\end{figure}

The first method is based on the bootstrap, a commonly used technique for constructing null distributions by random resampling with replacement. From the observed data $(\mathbf{x},\mathbf{y})$,
we first find the preferred nonlinear SEM, i.e. the one with a greater $\bar{R}^2$ statistic, and its estimated piecewise function $\hat{f}$ and the associated parameters. Next, we modify our data according to the null hypothesis before resampling. This is best illustrated with an example. For a dataset $(\mathbf{x},\mathbf{y})$ shown in Figure~\ref{fig:hypothesis_test_a}, the preferred model is $x \to y$ and the fitted function is represented by two red solid lines. We divide the data into two segments by the estimated cut point $\hat{\tau}_x$, indicated by the red dashed line in the figure. Then, we move one segment of the data points up or down along the $y$-axis by a minimum distance such that the two fitted line segments do not overlap in the range of the data. As $\hat{f}(x)$ after this modification becomes invertible, this leads to a modified data set $(\mathbf{x}_0,\mathbf{y}_0)$ (Figure~\ref{fig:hypothesis_test_b}) that satisfies the null hypothesis. As confirmed in Figure~\ref{fig:hypothesis_test_c} and \ref{fig:hypothesis_test_d}, the modified data $(\mathbf{x}_0,\mathbf{y}_0)$ can be fitted with an invertible nonlinear function, and the model fitting is comparable between the two directions $x_0\to y_0$ in panel \ref{fig:hypothesis_test_c} and $y_0\to x_0$ in panel \ref{fig:hypothesis_test_d}.

After generating the modified data $(\mathbf{x}_0,\mathbf{y}_0)$, our bootstrap testing procedure works as follows: 
\begin{enumerate}
\item Sample the null data $(\mathbf{x}_0,\mathbf{y}_0)$ with replacement to generate a bootstrap sample $(\mathbf{x}_0^b,\mathbf{y}_0^b)$;
\item For a bootstrap sample $(\mathbf{x}_0^b,\mathbf{y}_0^b)$, fit two nonlinear SEMs, one for each of the two directions;
\item Calculate the comparison statistic between the two directions, $\eta_{0}^b$ using equation \eqref{eq:test_stat}; 
\item Repeat the first three steps for $b=1,\ldots,B$ to generate the bootstrap null distribution of $\{\eta_0^b\}$, based on which we can calculate the $p$-value of $\hat{\eta}$ in our test.
\end{enumerate}

\begin{figure}
    \centering
    \includegraphics[width=0.5 \linewidth]{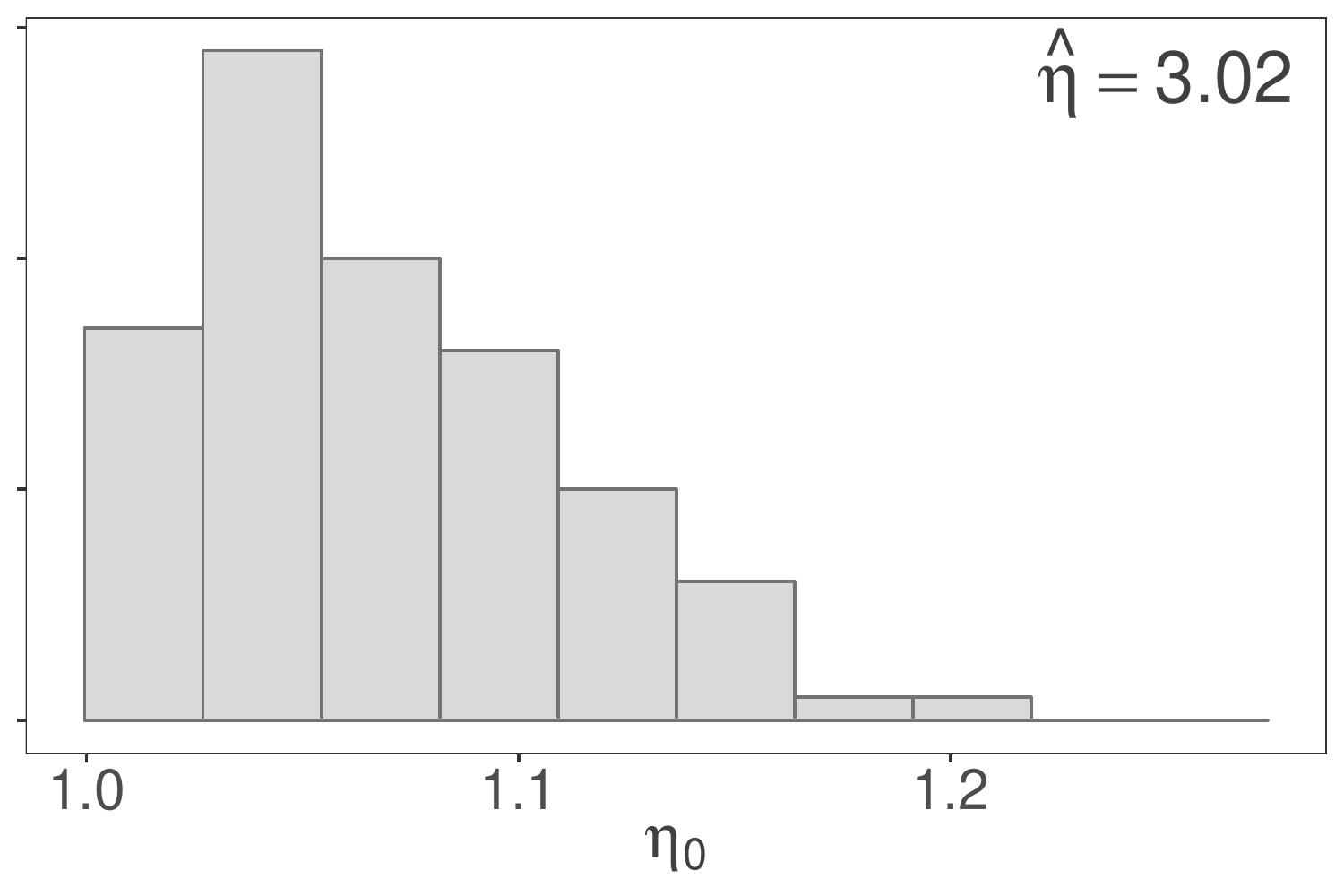} 
    \caption{Null distribution of $\eta_0$ estimated by the histogram of $\{\eta_0^b\}$.} \label{fig:hist_h0}
\end{figure}

Figure~\ref{fig:hist_h0} shows the bootstrap distribution of $\eta_0$ from the example in Figure~\ref{fig:hypothesis_test}. It ranges from 1 to 1.2, while the observed $\hat{\eta}=3.02$ is way much larger, indicating that $f$ is not invertible as shown in Figure~\ref{fig:hypothesis_test}.

The bootstrap method can be computationally intensive, especially for approximating small $p$-values. Therefore, we develop a more efficient alternative method to approximate the null distribution of the test statistic $\eta$ and calculate the $p$-values. We know from the definition of $\eta$ that it is a function of Pearson's correlation coefficients. It is well-known that Pearson's correlation coefficient after Fisher transformation follows approximately a normal distribution when the sample size $n$ is large. Thus, the distribution of $\eta_0$ can be obtained by sampling Pearson's correlation coefficient from this approximate distribution. The details are presented below.

Let $r$ be the Pearson's correlation coefficient of $(\mathbf{x},\mathbf{y})$. Fisher's z-transformation of $r$ is
$$z = \frac{1}{2} \log \frac{1+r}{1-r} = \arctan(r).$$
If $(\mathbf{x},\mathbf{y})$ is an i.i.d. sample from a bivariate normal distribution with true correlation $\rho$, then $z$ is approximately normally distributed as $\mathcal{N}(\arctan(\rho), {1}/{(n-3)})$, where $n$ is the sample size.
Assuming that each segment of the null data $(\mathbf{x}_0,\mathbf{y}_0)$ follows a bivariate normal distribution, then the inverse of the transformation $r = \tanh{(z)}$ can be used to construct the distribution of $\eta_0$. 
Given the null data, we first estimate the optimal cut point in each direction (the red and blue dashed lines in Figure~\ref{fig:hypothesis_test_c}). For the direction $x_0 \to y_0$, we separate the null data into two subsets: $(\mathbf{x}_0,\mathbf{y}_0)_l$ and $(\mathbf{x}_0,\mathbf{y}_0)_h$ according to the estimated cut point $\hat{\tau}_{x_0}$ of $\mathbf{x}_0$. For each subset of data, we compute its correlation coefficient, denoted by $\rho_l, \rho_h$ respectively. Now sample $z_l, z_h$ from
$$z_{l} \sim \mathcal{N} \big(\arctan(\rho_l), {1}/{(n_{l}-3)} \big), z_{h} \sim \mathcal{N} \big( \arctan(\rho_{h}), {1}/{(n_{h}-3)} \big),$$
where $n_l, n_h$ are the sample sizes of $(\mathbf{x}_0,\mathbf{y}_0)_l$ and $(\mathbf{x}_0,\mathbf{y}_0)_h$, respectively.
Substituting Pearson's correlation coefficient $r$ with $\tanh(z)$ in the formula of $\bar{R}^2$ in Equation~\eqref{eq:R2}, we get
\begin{align*}
    \bar{R}^{2}_{x_0 \to y_0} = [n_{l} \tanh(z_{l})^2 + n_{h} \tanh(z_{h})^2] / (n_{l} + n_{h}).
\end{align*}
Similarly, we can draw $\bar{R}^{2}_{y_0 \to x_0}$ using the same procedure, and obtain a large sample of $\eta_0$ to approximate the null distribution.

\begin{figure}
    \centering
    \includegraphics[width=0.5\linewidth]{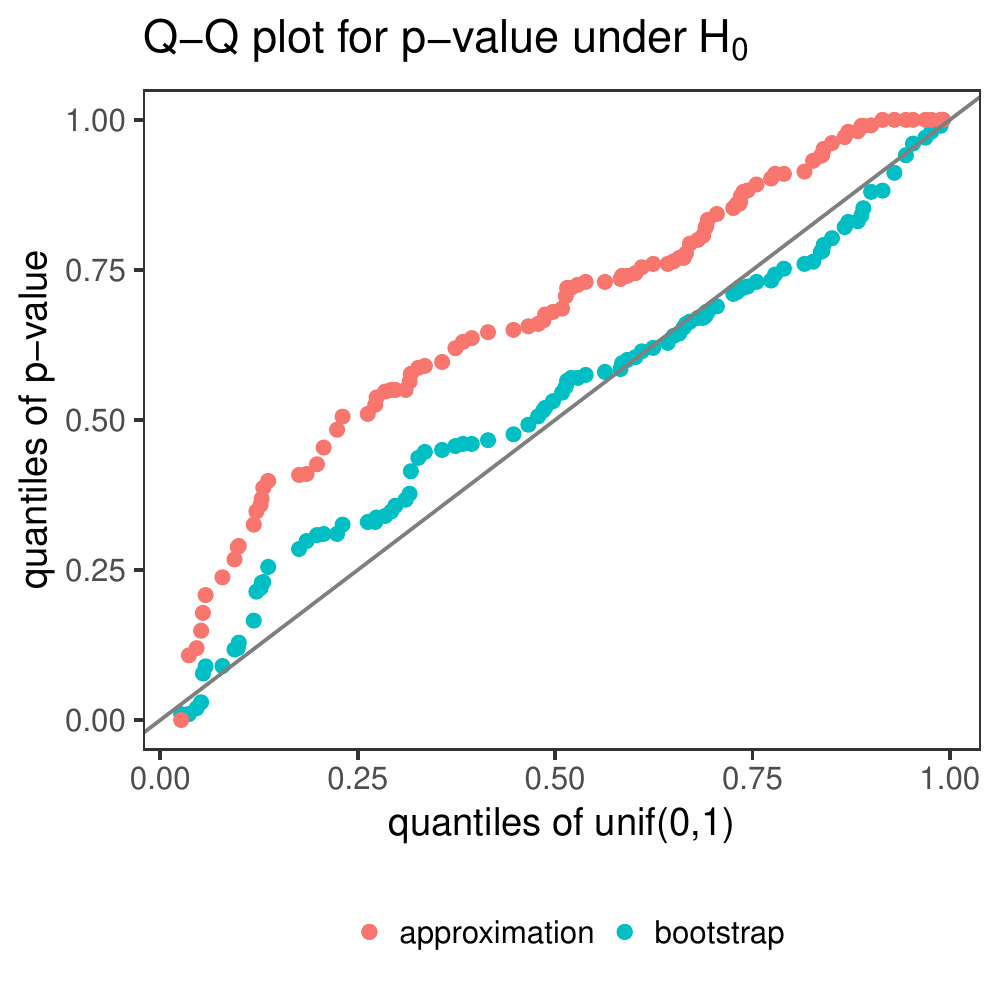}
    \caption{The Q-Q plot of $p$-value distributions under null against Unif(0,1).} \label{fig:qqplot_pvalue}
\end{figure}

Simulation was performed to validate $p$-values calculated by the bootstrap and the normal approximation procedures under the null hypothesis. We generated 100 data sets under invertible SEMs, and used the above two procedures to calculate the $p$-value for each dataset. Figure~\ref{fig:qqplot_pvalue} shows the quantile-quantile plots of these $p$-values against $\text{Unif}(0,1)$. The bootstrap $p$-values are approximately uniformly distributed between $(0, 1)$ while the $p$-values calculated via normal approximation seem to be a little left-skewed compared to the uniform distribution. Accordingly, at a significance level of $0.05$, the rejection rate was controlled at $0.05$ for the bootstrap test, while the normal approximation $p$-values were more conservative with a rejection rate around $0.02$. Note that for both tests, the type-I error was controlled at or below the desired level of $0.05$.

\subsection{Algorithm for bivariate case} \label{ss:bivariate_alg}

We summarize below our Algorithm~\ref{alg:bivariate} for inferring the causal relation between two variables from observed data $(\mathbf{x}, \mathbf{y})$. This algorithm will serve as a unit in our structure learning of causal networks in Section~\ref{sec:causal_structure_learning}. Thus, we represent its output as an edge between the two variables $X$ and $Y$, regarding as two nodes in a graph. There are four possible outcomes: $E(X,Y)\in\{\emptyset,X\to Y, Y\to X, X-Y\}$. The case $E(X,Y)=\emptyset$ means that the two variables are independent (no edge between the two nodes). An undirected edge $X-Y$ indicates that the SEM is invertible and the causal direction cannot be decided. A directed edge will be output if the test in the previous section is rejected at the significance level $\alpha$. 

\begin{algorithm}
\caption{Bivariate non-invertible causal discovery algorithm}
\begin{algorithmic}[1] \label{alg:bivariate}
\REQUIRE Observed data $(X, Y)$, significance level $\alpha$
\end{algorithmic}
\begin{algorithmic}[1]
\STATE Initialize $E(X,Y)=\emptyset$
\STATE Independence test between $X$ and $Y$ \label{alg_bivariate_ln:ind_test}
\IF{$X$ $\not\perp$ $Y$}
\STATE Fit a bivariate nonlinear SEM for each direction as described in Section~\ref{ss:bivariate_NISEM} and \ref{ss:model_fitting} \label{alg_bivariate_ln:model_fitting} 

\STATE Calculate corresponding goodness of fit statistics $\bar{R}_{X \to Y}^2, \bar{R}^2_{Y \to X}$ \label{alg_bivariate_ln:calculate_stats}
\STATE Preferred edge direction
$
E(X, Y) :=
\begin{cases}
X \to Y \quad \text{if} \quad \bar{R}_{X \to Y}^2 \geq \bar{R}^2_{Y \to X} \\
Y \to X \quad \text{if} \quad \bar{R}_{X \to Y}^2 < \bar{R}^2_{Y \to X} \\
\end{cases} 
$ \label{alg_bivariate_ln:edge_direction}

\STATE Calculate $p$-value of the causal direction test (Section~\ref{ss:hypothesis_testing}) \label{alg_bivariate_ln:pvalue} 
\STATE If $p>\alpha$, set $E(X,Y):=X-Y$
\ENDIF
\ENSURE
$E(X,Y)$  
\end{algorithmic}
\end{algorithm}

\section{Nonlinear causal structure learning} \label{sec:causal_structure_learning}

In this section, we incorporate non-invertible causal discovery into structure learning of a causal network among $p$ variables, $X_1,\ldots,X_p$. The generative distribution for these $p$ random variables is given by a set of SEMs whose structures are defined by an underlying directed acyclic graph (DAG), which will be called the causal DAG or causal graph. We allow both nonlinear and linear causal relations in the model. Our proposed method combines classic structure learning methods targeting at linear DAGs, such as the PC algorithm [\cite{Spirtes1991}] and regularized likelihood methods [\cite{Fu2013}], with our non-invertible causal discovery approach (Algorithm~\ref{alg:bivariate}).

Before a detailed description of our new method, we give a quick review of causal DAGs and general SEMs in Section~\ref{sec:DAGreview}.

\subsection{Causal DAGs and SEMs}\label{sec:DAGreview}

\cite{pearl2009} and \cite{Spirtes2010} pioneered the use of DAGs in causal modeling and inference. A causal DAG $\mathcal{G}$ on a set of variables $X_1,\ldots,X_p$ encodes assumptions about the data-generating process and is a great tool for visualization of causal relations among these variables. 
There is a directed edge $X_i \to X_j$ if and only $X_i$ is a direct cause of $X_j$, in which case we say $X_i$ is a (causal) parent of $X_j$. Let $\textbf{PA}_i$ denote the set of parents of $X_i$. The joint generative distribution $\mathbb{P}$ over the variables $\mathbf{V}=\{X_1,\ldots,X_p\}$ modeled by the causal DAG $\mathcal{G}$ is specified by a set of SEMs
\begin{align}\label{eq:genSEM}
X_i = f_i(\textbf{PA}_i, \epsilon_i), \quad i = 1, \ldots, p,
\end{align}
where $\epsilon_i$ is a background or noise variable independent of $\textbf{PA}_i$. Moreover, all background variables $\epsilon_j$, $j=1,\ldots,p$ are mutually independent so that the joint distribution $\mathbb{P}$ satisfies Markov properties with respect to the causal DAG $\mathcal{G}$.

We make the following assumptions on the above causal DAG model:
\begin{enumerate}
    \item \textbf{Causal sufficiency:} The set of variables $\mathbf{V}$ is \textit{causally sufficient}. That is, there is no variable not in $\mathbf{V}$ that is a direct cause of more than one variable in $\mathbf{V}$ [\cite{Spirtes2010}]. In other words, all common causes of variables in the DAG are included in the set $\mathbf{V}$ of measured variables.
    \item \textbf{Faithfulness:} Every conditional independence relation implied by the joint distribution $\mathbb{P}$ over $\mathbf{V}$ is entailed by d-separation in the causal DAG [\cite{Spirtes2010}].
\end{enumerate}
These are common assumptions in structure learning of DAGs [\cite{Spirtes1991, Chickering2003}]. 

\subsection{Restricted equivalence class} \label{ss:restricted_equi_class}

In this work, we assume that there are both linear and nonlinear causal relations in the SEMs~\eqref{eq:genSEM} under an additive model framework: 
\begin{align}\label{eq:lnSEM}
X_i = \sum_{j \in \textbf{PA}_i^l} {\beta_{ji}} X_j  + \sum_{k \in \textbf{PA}_i^{n}} f_{ki}(X_k) + \epsilon_i, \quad i = 1, ..., p,
\end{align}
where $f_{ki}$ is a nonlinear function. We call $\textbf{PA}_i^l$ and $\textbf{PA}_i^{n}$ the linear and nonlinear parent sets of $X_i$, and accordingly, call an edge $j\to i$ a linear and nonlinear edge, respectively, for $j\in\textbf{PA}_i^l$ and $j\in\textbf{PA}_i^{n}$.  

\begin{figure}
    \centering
    \includegraphics[width= 0.8\linewidth]{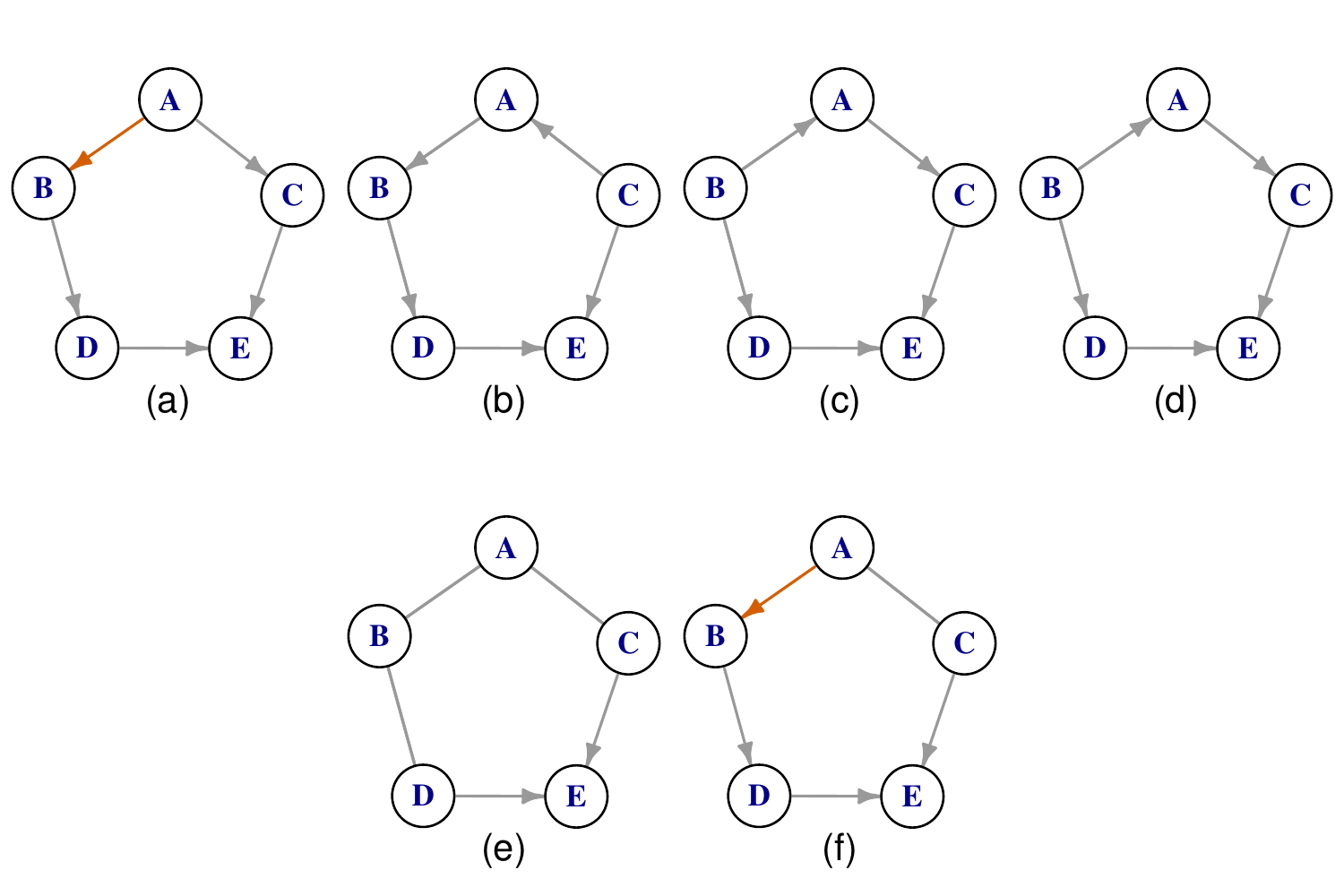}
    \caption{DAGs (a) to (d) in an equivalence class represented by the CPDAG (e), and the restricted CPDAG (f) subject to a non-invertible edge (the red edge).}  
    \label{fig:causal_dag}  
\end{figure}

When $\textbf{PA}_i^n = \emptyset$ in Equation~\eqref{eq:lnSEM}, we have the regular linear SEMs. Linear SEMs with Gaussian errors are not identifiable due to the so-called Markov equivalence class of DAGs, which is a set of DAGs encoding the same set of conditional independence relations. Two DAGs are Markov equivalent if and only if they have the same skeleton and the same $v$-structures [\cite{Verma1990}]. Here, the skeleton of a DAG is the underlying undirected graph obtained by ignoring the direction of every edge, and a $v$-structure is an ordered triplet of nodes $(i,j,k)$ of the form $i\to k \leftarrow j$, where $i, j$ are not connected by an edge. 
A Markov equivalence class can be uniquely represented by a {completed partially DAG} ({CPDAG}), which contains both directed and undirected edges. As illustrated in Figure~\ref{fig:causal_dag}, DAGs (a)\textendash(d) have the same skeleton and the same $v$-structure $C\to E \leftarrow D$, and they constitute all the DAGs in the Markov equivalence class, which is represented by the corresponding CPDAG (e).

If some of the undirected edges in a CPDAG can be oriented, say by non-invertible relations in our problem, then the equivalence class will be reduced. One can apply Meek's rules [\cite{Meek1995}] to orient other undirected edges and obtain a maximally oriented partially DAG (PDAG) that represents a restricted equivalence class. We call this maximally oriented PDAG a \textit{restricted CPDAG}, which serves as the ground-truth for our structure learning. Suppose the edge $A \to B$ of the DAG (a) in Figure~\ref{fig:causal_dag} is non-invertible and thus not reversible. Keeping the orientation of this edge in the CPDAG (e), we then maximally orient the rest of the undirected edges, which leads to the orientation of $B \to D$ since $D\to B$ would introduce an extra $v$-structure with the non-reversible edge $A \to B$. Thus we obtain the restricted CPDAG (f) for this example. In general, a restricted CPDAG, subject to a set of non-reversible edges, represents the subset of DAGs in the Markov equivalence class that have the same orientations for those non-reversible edges. In Figure~\ref{fig:causal_dag}, the restricted equivalence class includes DAGs (a) and (b), out of the four DAGs in the Markov equivalence class. 


\subsection{Structure learning algorithms} \label{ss:algorithms}

Our goal is to infer the causal DAG, with both linear and nonlinear edges, from observational data. 
The overall idea of our approach is to combine an existing linear structure learning algorithm to identify a CPDAG from the data. Then we recursively apply the bivariate non-invertible causal discovery algorithm (Algorithm~\ref{alg:bivariate}) in Section~\ref{ss:bivariate_alg} to detect any non-invertible relation and orient more edges. Since linear structure learning algorithms may output a DAG, a CPDAG or a PDAG, we in general assume the output is a PDAG which includes CPDAGs as a special case. Note that a DAG learned by linear structure learning will be converted to a CPDAG before applying our algorithm.
Given a PDAG, we first develop a non-invertible nonlinear causal learning (NNCL) algorithm that generalizes the bivariate algorithm described in Section~\ref{ss:bivariate_alg} to multiple variables. Then we discuss a few approaches that combine a linear structure learning algorithm with the NNCL algorithm to identify a causal graph with both linear and nonlinear edges.

We distinguish directed and undirected neighbors in a PDAG as follows. If there is a directed edge $j\to i$ in a PDAG, we say $j$ is a parent of $i$ and if they are linked by an undirected edge $i-j$, they are called a neighbor of each other. We define $\textbf{PA} = \{ \textbf{PA}_i, i = 1, ... p \}$ as the collection of all parent sets and $\textbf{U} = \{(X_i, X_j): i-j \in E\}$ as the set of all undirected edges in a PDAG $\mathcal{G}=(\mathbf{V},E)$, where $\mathbf{V}$ is the node set and $E$ is the edge set.

Based on an input initial PDAG $\mathcal{G}$, our NNCL algorithm recursively detects the most significant non-invertible edge among all undirected ones, then fixes the orientation of this edge in the graph and applies the orientation rules [\cite{Meek1995}] to orient the remaining undirected edges. A non-invertible edge between $X_i$ and $X_j$ is detected by reducing to the bivariate case (Algorithm~\ref{alg:bivariate}) after calculating the residuals after projecting each of them to its respective identified parents.
An outline of our algorithm is shown in Algorithm~\ref{alg:NNCL}. 

\begin{algorithm}
\SetAlgoLined
\caption{Non-invertible nonlinear causal learning (NNCL)}
\begin{algorithmic}[1]\label{alg:NNCL}
\REQUIRE observed data for $(X_1, ..., X_p)$, initial PDAG $\mathcal{G}=(\mathbf{V},E)$, significance level $\alpha$
\medskip
\REPEAT
\FOR{$(X_i, X_j) \in \textbf{U}$}
\STATE apply Algorithm~\ref{alg:bivariate}\label{alg_nncl_ln:model_fitting}  Line~\ref{alg_bivariate_ln:model_fitting}-\ref{alg_bivariate_ln:pvalue} on the residuals of $X_i$, $X_j$ after regressing on their respective parents $\textbf{PA}_i, \textbf{PA}_j$
\STATE \textbf{obtain:} test-statistic $\eta_{i,j}$, $p$-value $p_{i,j}$, and the preferred direction $E_{i,j}$ 
\ENDFOR

\STATE Sort $\textbf{U}$ by $p_{i,j}$ $^*$, and let $E_{(1)}=(V_p\to V_c)$ be the edge with minimum $p$-value $p_{(1)}$. 

\IF{\begin{enumerate}[label=(\roman*)]
    \item $p_{(1)} \leq \alpha$
    \item $V_p \not\perp V_c | \textbf{PA}_{V_c}$
    \item adding $E_{(1)}$ to $\mathcal{G}$ does not induce any directed cycle
\end{enumerate} 
}\label{alg_nncl_ln:conditions}
\STATE add $E_{(1)}$ to $E$
\STATE orient other undirected edges in $\mathcal{G}$ by Meek's rules \label{line:meek}
\STATE update $\{ \textbf{U}, \textbf{PA}\}$
\ENDIF
\UNTIL no more edge can be added

\ENSURE Restricted CPDAG $\mathcal{G}$
\end{algorithmic}
$*$ when there is a tie, sort by $\eta_{i,j}$
\end{algorithm}

The initial residuals in Line~\ref{alg_nncl_ln:model_fitting} are calculated from regressing $X_i$ and $X_j$ on their respective linear parents in the initial PDAG $\mathcal{G}$. Then in the following steps, every time a nonlinear parent is added to the structure, the residuals of the child node will be updated to the residuals calculated from the fitted piecewise function. Note that Line~\ref{alg_nncl_ln:conditions}(ii) is used in place of the independence test (Line~\ref{alg_bivariate_ln:ind_test}) in Algorithm~\ref{alg:bivariate}. For a preferred edge $V_p \to V_c$, where $V_p,V_c\in\mathbf{V}$, we first divide the data according to the cut point $\hat{\tau}$ of $V_p$ estimated in the piecewise linear function, and then perform conditional independence test for each segment of the data. We require both reject the null hypothesis in order to conclude that $V_p \not\perp V_c | \textbf{PA}_{V_c}$. This procedure takes into account the nonlinear relationship between $V_p$ and $V_c$. 
See supplementary material for the details on the conditional independence tests in our procedure.

The initial PDAG in Algorithm~\ref{alg:NNCL} can be estimated using an existing structure learning algorithm that produces a CPDAG from observational data. However, the initial graph may fail to detect the dependency among variables in a nonlinear relationship, thus missing nonlinear edges in the estimated skeleton. Therefore, we implement the following algorithm to search outside the skeleton of the initial PDAG after Algorithm~\ref{alg:NNCL} is done.

\begin{algorithm}
\caption{Searching nonlinear edges outside the initial skeleton}
\begin{algorithmic}[1]\label{alg:outside_search}
\REQUIRE observed data for $(X_1, ..., X_p)$, $\mathcal{G}$ output from Algorithm~\ref{alg:NNCL}, significance level $\alpha$
\medskip

Define the set of non-adjacent pairs in $\mathcal{G}$ as $\textbf{NE}$ 
\FOR{$(X_i, X_j) \in \textbf{NE}$} 
\STATE  apply Algorithm~\ref{alg:bivariate} Line~\ref{alg_bivariate_ln:model_fitting}-\ref{alg_bivariate_ln:pvalue} on the residuals of $X_i, X_j$ after regressing on their respective parents $\textbf{PA}_i, \textbf{PA}_j$
\STATE \textbf{obtain:} $p$-value $p_{i,j}$, and the preferred direction $E_{i,j}=V_p\to V_c$ 
\IF{ 
\begin{enumerate}[label=(\roman*)]
    \item $p_{{i,j}} \leq \alpha$
    \item $V_p \not\perp V_c | \textbf{PA}_{V_c}$
    \item adding $E_{i,j}$ to $\mathcal{G}$ does not induce any directed cycle
\end{enumerate}
}
\STATE add $E_{i,j}$ to $E$ 
\STATE orient other undirected edges in $\mathcal{G}$ by Meek's rules
\ENDIF
\ENDFOR
\ENSURE Restricted CPDAG $\mathcal{G}$
\end{algorithmic}
\end{algorithm}

In our implementation, we use two structure learning algorithms to construct the initial estimate of a CPDAG: the order-independent PC algorithm [\cite{colombo14}] and the CCDr algorithm [\cite{aragam2015}].
The PC algorithm is a constraint-based method that learns a graphical structure by repeated conditional independence (CI) tests. The main procedure of this method is to first estimate a skeleton using CI tests, and then identify $v$-structures in the skeleton. Finally it applies the orientation rules in \cite{Meek1995} to direct the remaining edges without introducing new conditional independence relations or directed cycles. The PC algorithm we use is implemented in the \textit{bnlearn} package [\cite{Scutari2009}], and the details of the algorithm can be found in \cite{colombo14}.
The CCDr algorithm is a score-based method that maximizes a regularized likelihood under a concave penalty function. This algorithm is available in the R package \textit{sparsebn} [\cite{aragam2019}] with algorithm details described in \cite{aragam2015}. The CCDr algorithm outputs a DAG, which we convert to a CPDAG by the function \textit{cpdag} in the package \textit{bnlearn}. We call these two implementations PC-NNCL and CCDr-NNCL, respectively.



We discuss briefly the intuition behind our algorithms. Assume that 
\begin{enumerate*}[label=\arabic*)]
    \item the input initial PDAG $\mathcal{G}$ in Algorithm~\ref{alg:NNCL} is the CPDAG of the true DAG;
    \item there exists an undirected edge between $X_i$ and $X_j$ that is non-invertible. 
\end{enumerate*}
Algorithm~\ref{alg:NNCL} will first obtain residuals by regressing $X_i$ and $X_j$ on their respective identified parents $\textbf{PA}_i$ and $\textbf{PA}_j$. Since we assume that the parent effects are additive, the residuals will preserve a non-invertible relation, which reduces the problem to the bivariate case. 
Thus, we will be able to direct this non-invertible edge using Algorithm~\ref{alg:NNCL} when the sample size becomes large. After that, a repeated application of Meek's orientation rules (Line~\ref{line:meek}) will maximally orient the graph and recover the restricted CPDAG. It is possible that the initial PDAG $\mathcal{G}$ may not contain certain non-invertible edges. The additional search procedure in Algorithm~\ref{alg:outside_search} is designed to detect such missing edges. 


\section{Numerical experiments} \label{sec:experiments}

In this section, we report numerical experiments on simulated data to verify the validity and demonstrate the effectiveness of our non-invertible nonlinear causal network learning method. 
We evaluate three different versions of our method: PC-NNCL and CCDr-NNCL, discussed above, and NNCL (Algorithm~\ref{alg:outside_search} with empty initial graph), so that we can see the usefulness of combining linear structure learning (PC and CCDr) with our nonlinear edge detection. We also compare them with PC and CCDr in Section~\ref{ss:simulation}, and another recent method for nonlinear DAG learning in Section~\ref{ss:ANM_comparison}.

\subsection{Simulation study} \label{ss:simulation}


We selected six different DAGs with graph size ranging from small to large from the Bayesian network repository (\url{http://www.bnlearn.com/bnrepository/}). For each network we simulated data with different percentages of nonlinear edges: $0\%$, $20\%$, $40\%$, $60\%$, $80\%$, $100\%$. Under each setting we generated 10 simulated data sets with sample size $n=1,000$. The nonlinear relationships were simulated using different quadratic functions and the errors were simulated from $\mathcal{N}(0, 1)$. 

The ground-truth we compare against is the true restricted CPDAG. We used the function \textit{cpdag} in \textit{bnlearn} to transform the true DAG to its restricted CPDAG by specifying a white list of non-invertible edges in the DAG.

The Structural Hamming Distance (SHD) and Jaccard Index (JI) are used to evaluate the performance of the algorithms. 
The SHD measures the difference between the estimated graphs and the true graph. It is defined as the number of edge additions, deletions or orientation corrections in order to match two PDAGs. Here, orientation corrections include reversal of a directed edge and a change from a directed edge to an undirected one and vice versa. Thus, a lower SHD indicates a better performance. JI measures the similarity between two graphs. It is the percentage of correct edges among the union of edges in the estimated graph and the true graph. The higher the percentage is, the closer the two graphs are. 

In the following simulation results, the test for causal direction was performed using the normal approximation approach with a $p$-value threshold of $0.01$. The significance level of the conditional independence tests in Algorithm~\ref{alg:NNCL}  and \ref{alg:outside_search} was set at $0.01$. For conditional independence tests in the PC algorithm, we used the predefined \textit{gaussCItest} with significance level $0.01$. The default settings were used for running the CCDr algorithm.

\begin{figure}
    \centering
    \captionsetup{justification=centering}
    \includegraphics[width=0.75\linewidth]{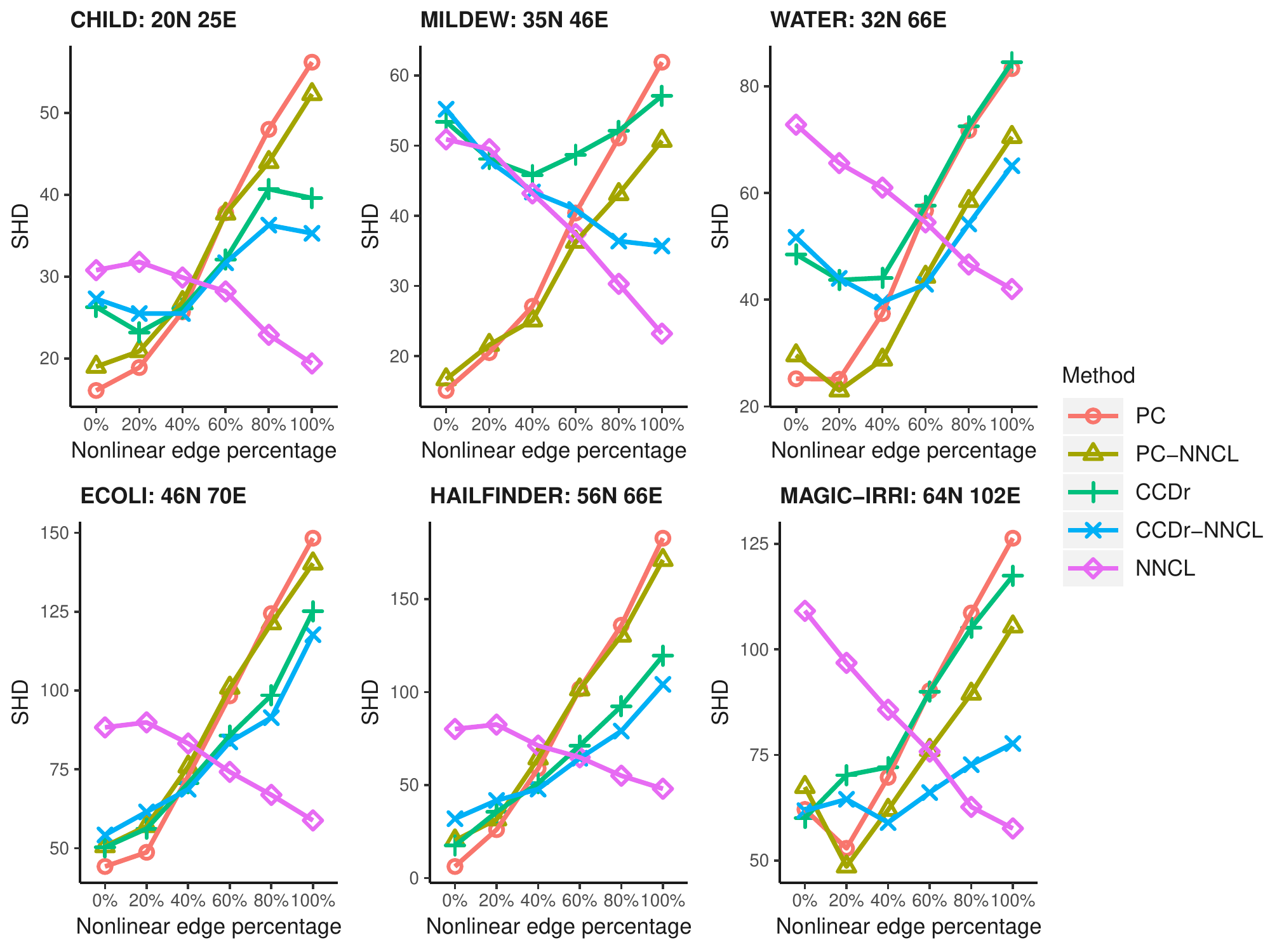}
    \caption{SHD comparison among five algorithms on six networks.\\ (N: number of nodes, E: number of edges)}
    \label{fig:networkSHD}
\end{figure}

\begin{figure}
    \centering
    \captionsetup{justification=centering}
    \includegraphics[width=0.75\linewidth]{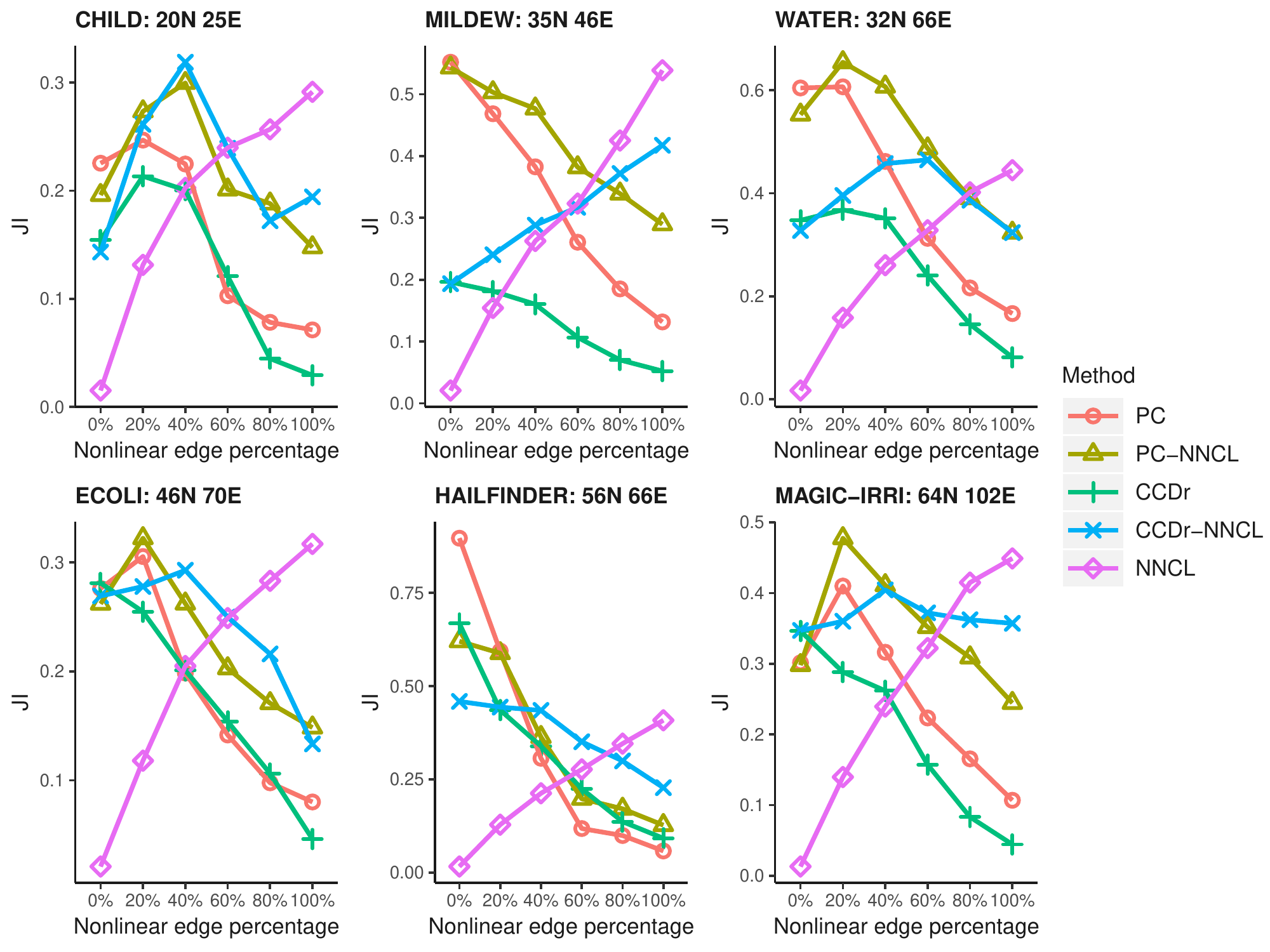}
    \caption{JI comparison among five algorithms on six networks.\\ (N: number of nodes, E: number of edges)}
    \label{fig:networkJI}
\end{figure}

Figures~\ref{fig:networkSHD} and \ref{fig:networkJI} show the SHD and JI comparisons among the five methods: PC, PC-NNCL, CCDr, CCDr-NNCL, and NNCL. The six panels in each figure report the results for the six networks from the Bayesian network repository. The colored curves correspond to different algorithms and are plotted against the percentage of nonlinear edges in the true DAG. 

PC and CCDr showed higher accuracy when there were less than $40\%$ nonlinear edges. As we increased the percentage of nonlinear edges, there was a dramatic decrease in the accuracies of PC and CCDr estimates, reflected by both metrics. This demonstrates the difficulty of these baseline algorithms in learning nonlinear DAGs. The NNCL algorithm exhibited an opposite trend, having higher accuracy for DAGs with more nonlinear edges. The SHDs of NNCL estimates were generally the smallest when there were more than $60\%$ of nonlinear edges. Linear-NNCL algorithms (i.e. PC-NNCL and CCDr-NNCL) showed great improvement over PC and CCDr, and the improvement became more substantial in settings with a higher percentage of nonlinear edges. The performance curves of the two linear-NNCL algorithms had a similar trend with the corresponding linear structure learning algorithms, because the nonlinear edge estimation was based on the initial graphs estimated by the linear algorithms. We observed moderate decrease in SHDs and significant increase in JI after the NNCL step, showing that more edges were correctly identified in the NNCL step. Overall, linear-NNCL algorithms showed the best performance for a wide range of nonlinear edge percentages. They became inferior to NNCL only when the true DAG was mostly composed of nonlinear edges ($\ge 80\%$).

\begin{figure}[h]
    \centering
    \includegraphics[width=0.7\linewidth]{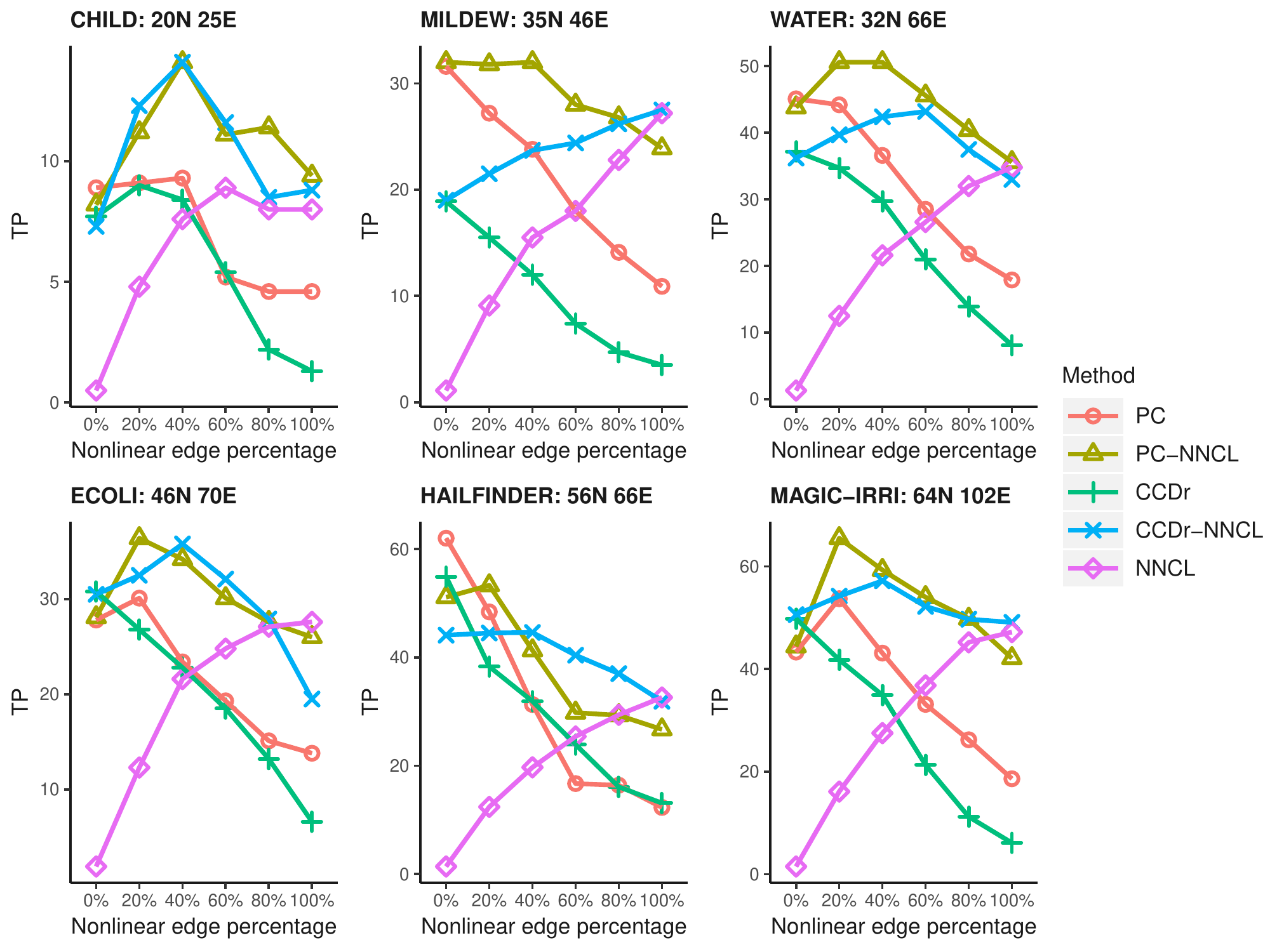}
    \caption{True positive comparison of five algorithms on six networks. (N: number of nodes, E: number of edges)}
    \label{fig_supp:networkTP}
\end{figure}

\begin{figure}[ht]
    \centering
    \includegraphics[width=0.7\linewidth]{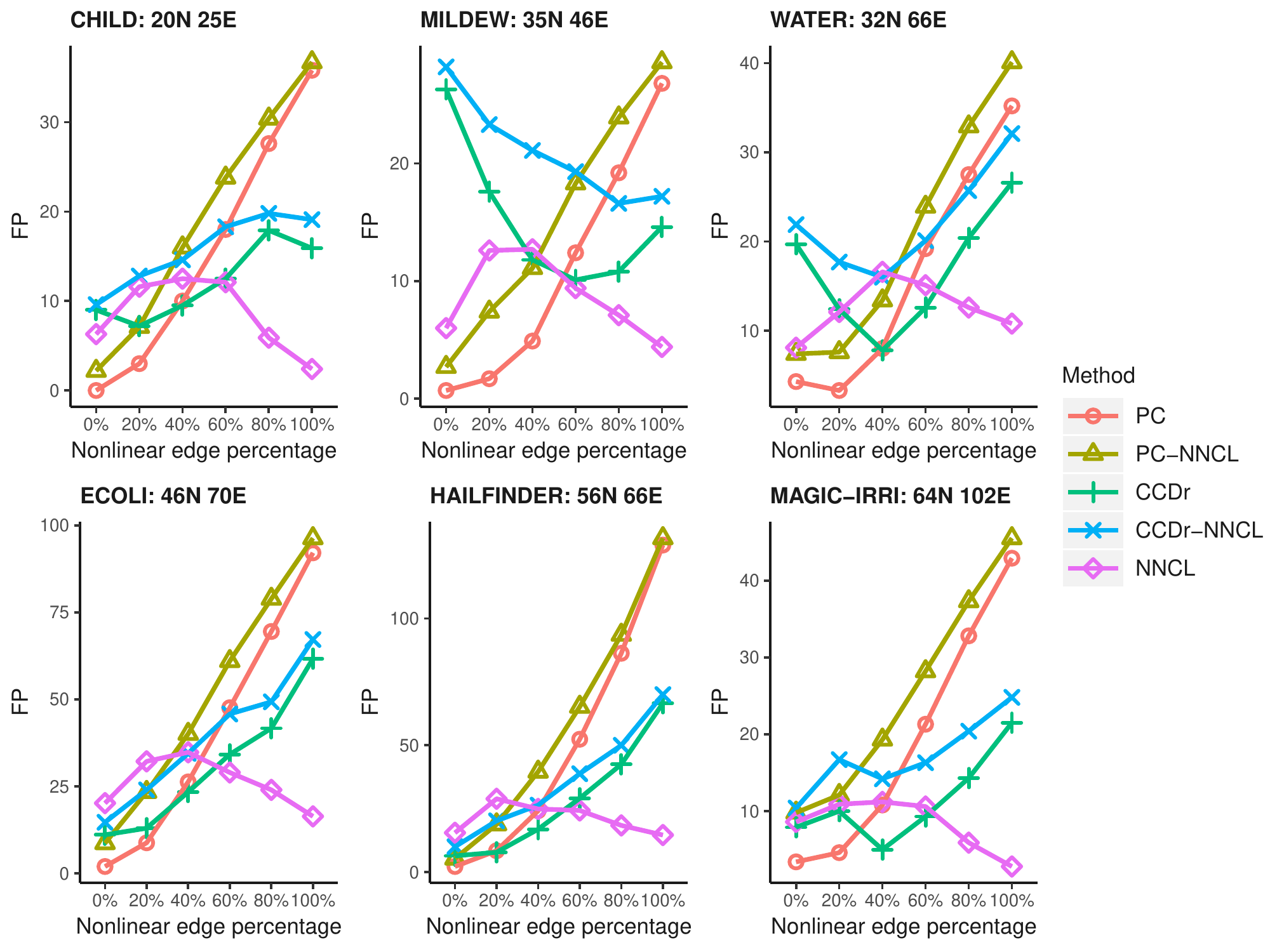}
    \caption{False positive comparison of five algorithms on six networks. (N: number of nodes, E: number of edges)}
    \label{fig_supp:networkFP}
\end{figure}

Besides the above overall accuracy metrics, we also report in the supplementary material the numbers of true positive (TP) and false positive (FP) edges in this comparison (Figure~\ref{fig_supp:networkTP}, \ref{fig_supp:networkFP}). 
True positive curves were similar to what we observed with the JI curves above. Adding NNCL step increased the TP by $15.2\%$ on average compared to the PC algorithm, and $20.8 \%$ on average compared to the CCDr algorithm. The NNCL algorithm had the lowest FP among all algorithms when there were more than $40\%$ nonlinear edges. The FP curves of linear-NNCL algorithms were close to those of the linear algorithms. Note that the NNCL step orients undirected edges and does not delete any edges in the initial CPDAG estimated by linear structure learning. Therefore, a linear-NNCL algorithm will not decrease the FP compared to its linear counterpart.

\begin{figure}[ht]
    \centering
    \includegraphics[width=0.7\linewidth]{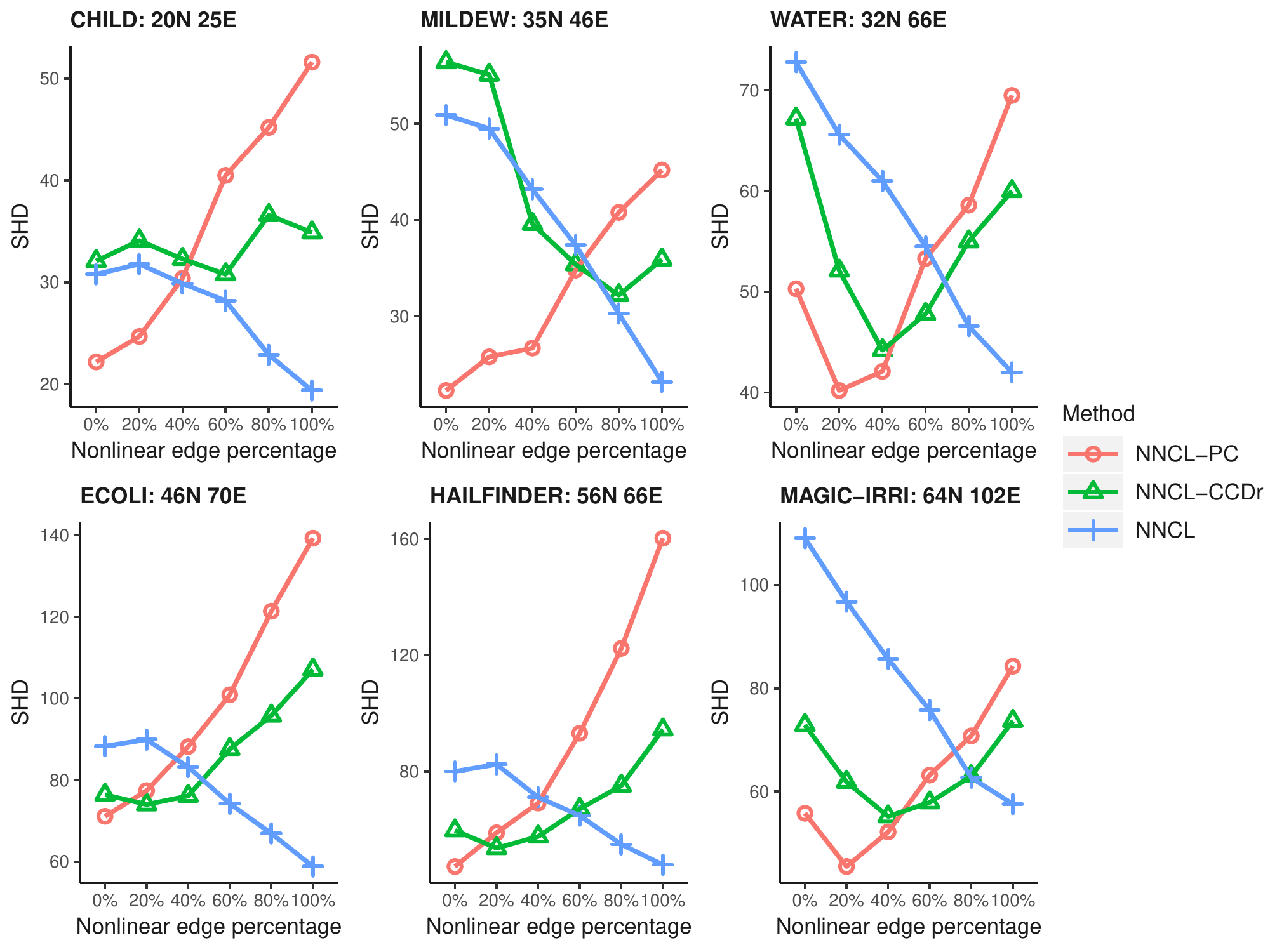}
    \caption{SHD comparison of three algorithms on six networks. (N: number of nodes, E: number of edges)}
    \label{fig_supp:networkSHD_alt_alg}
\end{figure}

\begin{figure}[ht]
    \centering
    \includegraphics[width=0.7\linewidth]{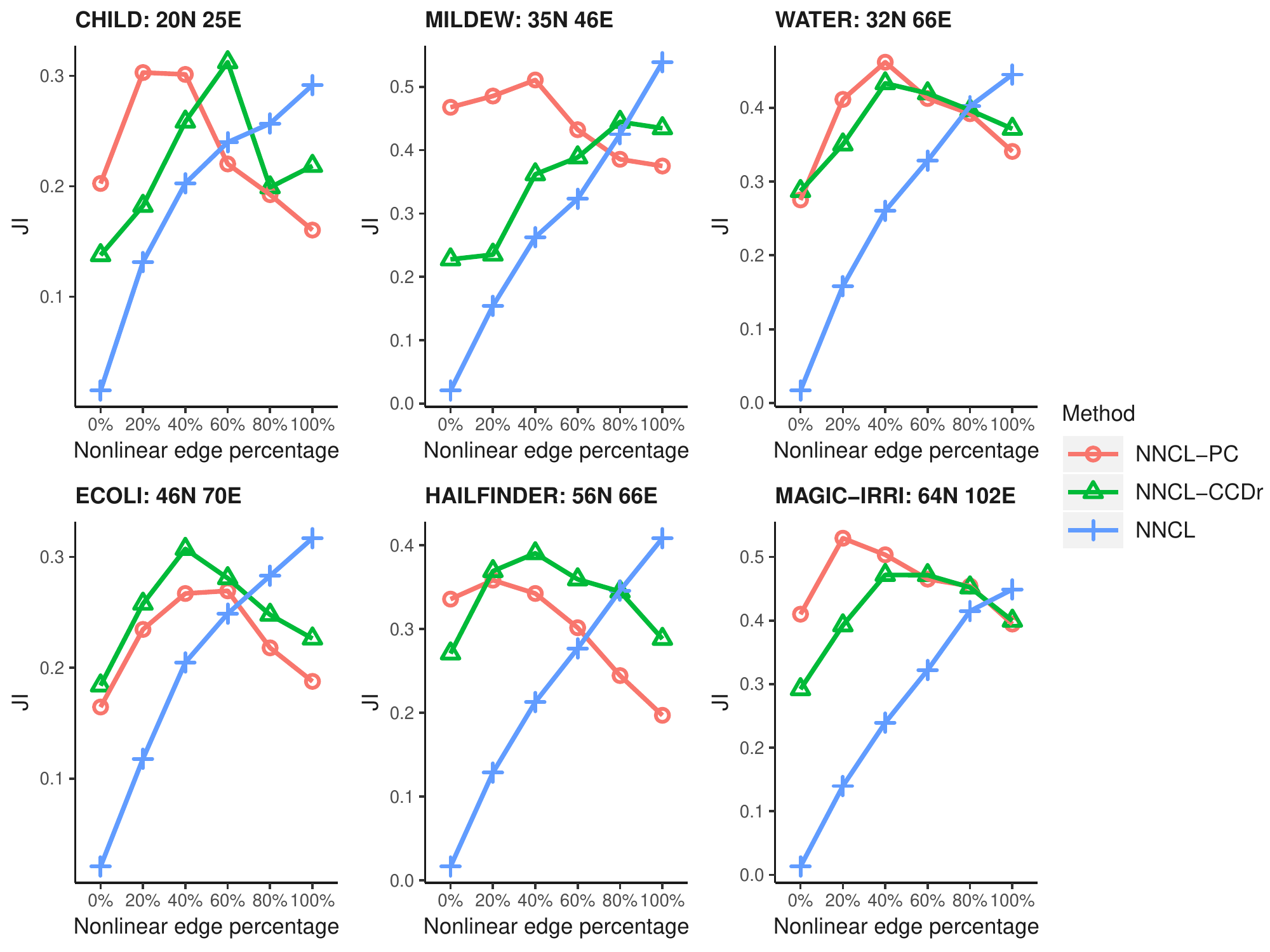}
    \caption{JI comparison comparison of three algorithms on six networks. (N: number of nodes, E: number of edges)}
    \label{fig_supp:networkJI_alt_alg}
\end{figure}

\begin{figure}[ht]
    \centering
    \includegraphics[width=0.7\linewidth]{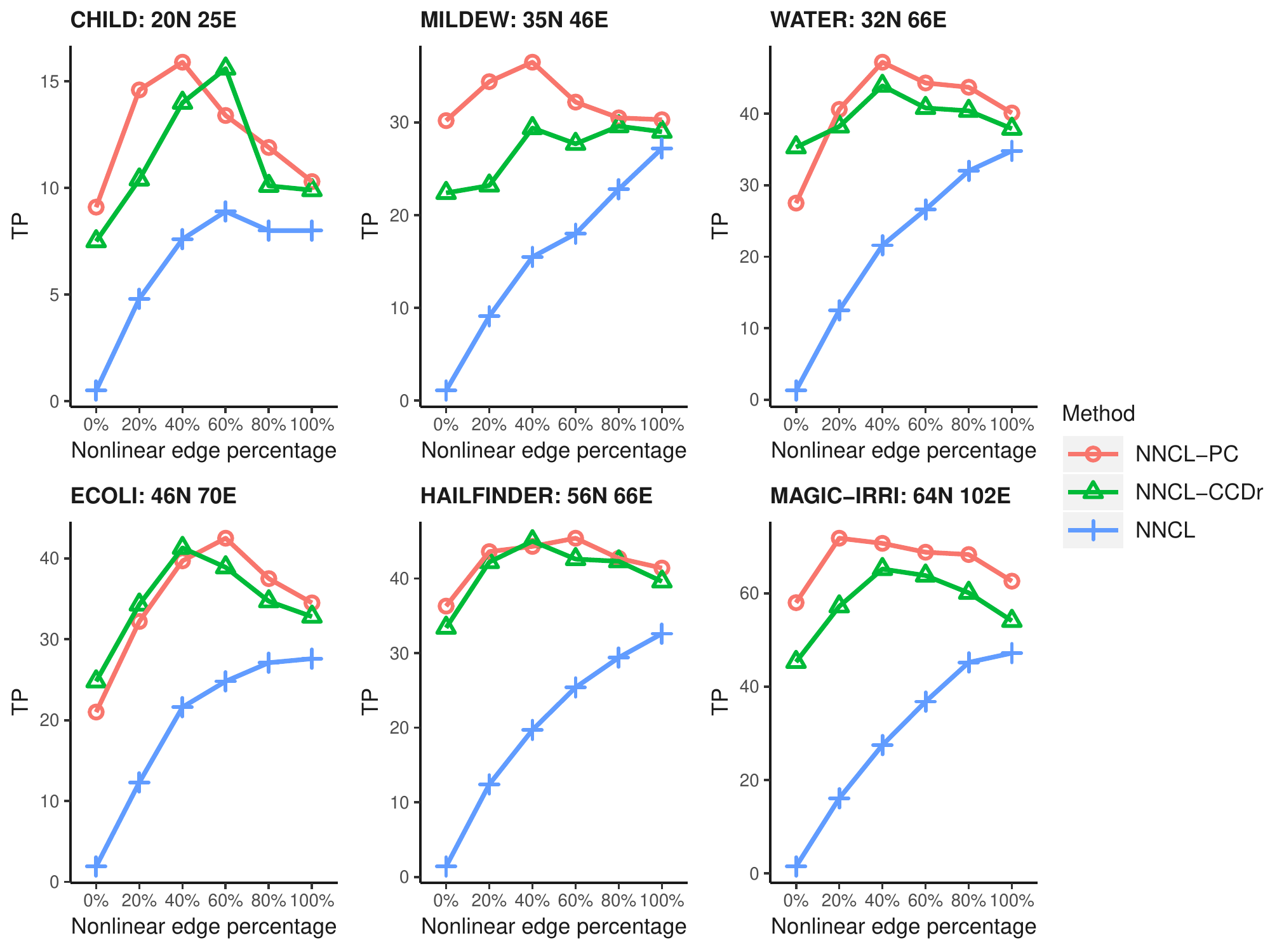}
    \caption{True positive comparison of three algorithms on six networks. (N: number of nodes, E: number of edges)}
    \label{fig_supp:networkTP_alt_alg}
\end{figure}

\begin{figure}[ht]
    \centering
    \includegraphics[width=0.7\linewidth]{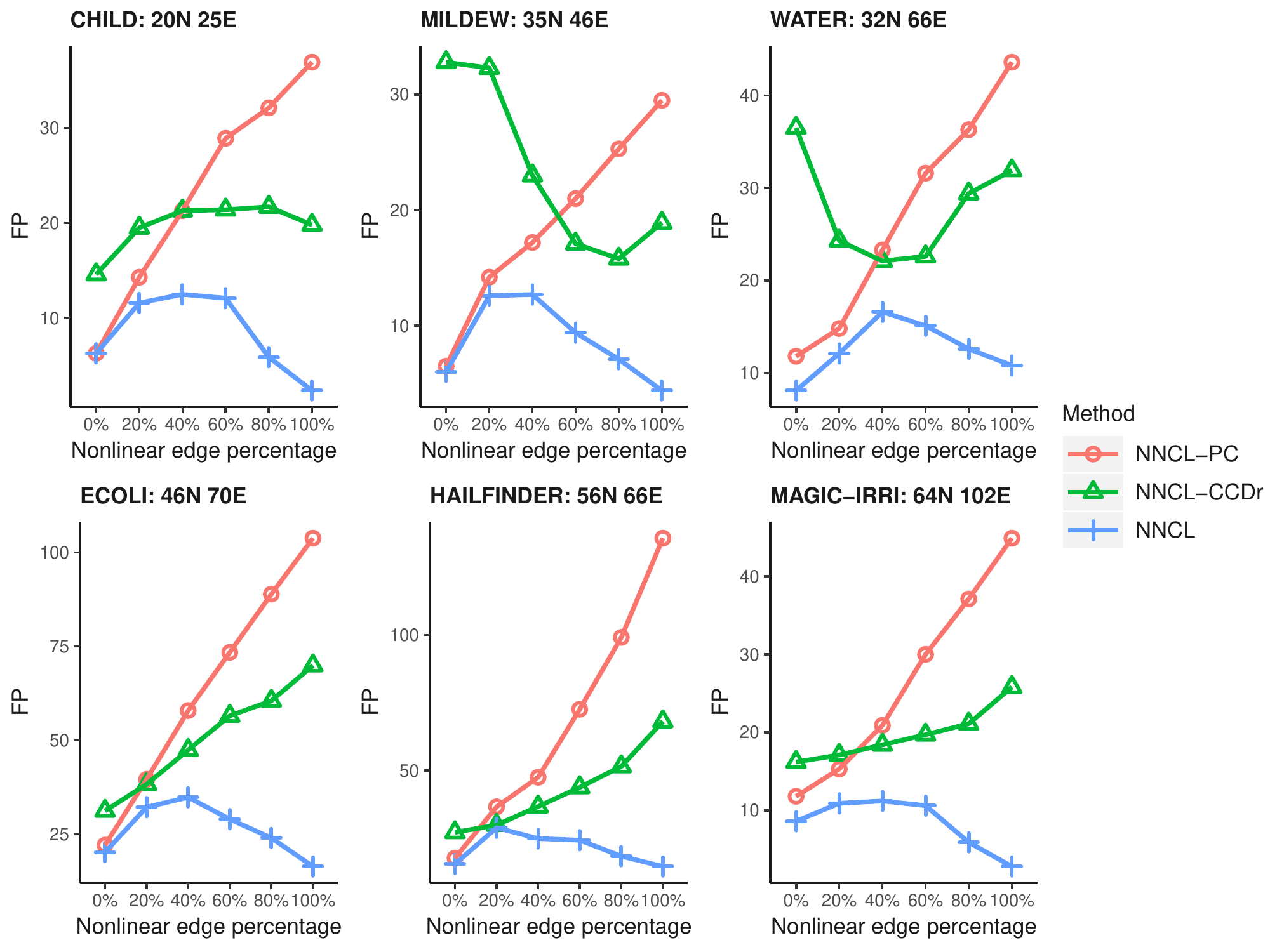}
    \caption{False positive comparison of three algorithms on six networks. (N: number of nodes, E: number of edges)}
    \label{fig_supp:networkFP_alt_alg}
\end{figure}

An alternative approach to learning a causal DAG with nonlinear edges is to first exhaustively search for nonlinear edges among all pairs of nodes by running Algorithm~\ref{alg:bivariate} repeatedly. Then we apply a linear structure learning algorithm with the detected nonlinear edges fixed. We call this approach NNCL-linear and present the results in the supplementary material.
The curves of JI and TP (Figure~\ref{fig_supp:networkJI_alt_alg} and \ref{fig_supp:networkTP_alt_alg}) show that, in general, adding the linear step after NNCL helped improve the detection of true positive edges, especially in settings with a low  percentage of nonlinear edges. In Figure~\ref{fig_supp:networkSHD_alt_alg}, we observe moderate decrease in SHDs of the NNCL-linear algorithms when there were less than $60\%$ nonlinear edges. However, we also observed a quite significant increase in SHDs of these algorithms comparing to NNCL when we increased the percentage of nonlinear edges to more than $60\%$. This is mainly due to the substantial increase in the FP edges in the linear step, especially for PC (see Figure~\ref{fig_supp:networkFP_alt_alg}). These observations suggest that when the DAG consists of mostly nonlinear edges, adding the linear step would be of little use. Overall, we found linear-NNCL algorithms more accurate and will stick to this approach in the following results.

\subsection{Comparison with RESIT} \label{ss:ANM_comparison}

Next, we compare our approach with a recent nonlinear causal learning algorithm called regression with subsequent independence test (RESIT) proposed by \cite{Peters2014}. RESIT was developed based on additive noise models (ANM)  [\cite{Hoyer2009}], which is identifiable from observational data. The algorithm consists of two phases. The first phase yields a topological ordering by iteratively identifying and removing a sink node. 
In each step of this iterative procedure, each of the remaining variables is regressed on all the other remaining variables and the dependence between residuals and the other variables is measured. The variable with the least dependence is identified as a sink node and removed. In the second phase, given the estimated ordering, superfluous edges are removed by further conditional independence tests. 
More details of the algorithm can be found in \cite{Peters2014}.

\begin{figure}
    \centering
    \includegraphics[width=0.5\linewidth]{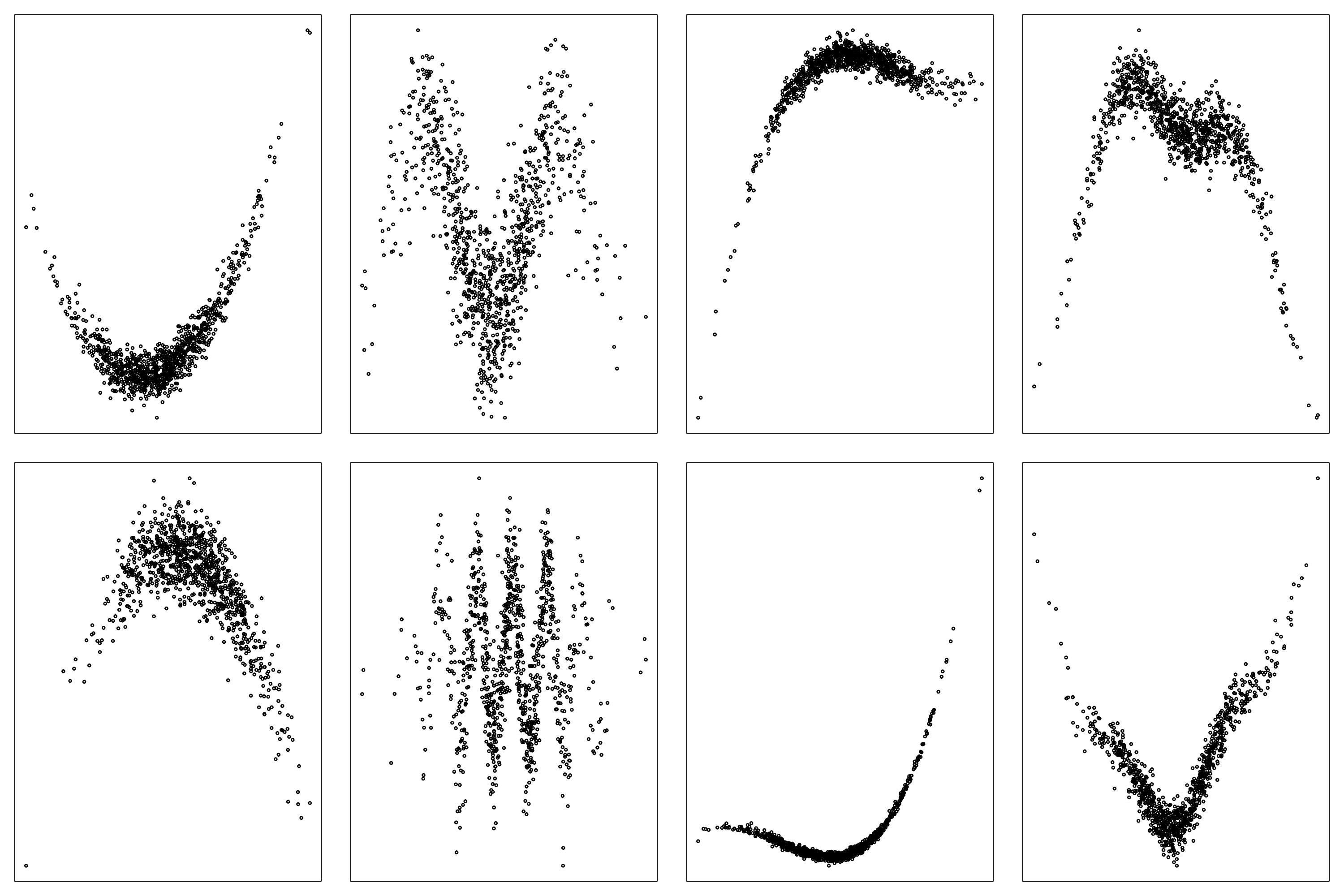}
    \caption{Examples of various nonlinear patterns in simulated data.}
    \label{fig:nonlinear_data_type}
\end{figure}

Since RESIT does not handle a large number of nodes effectively and generally takes a long time to run, we compared our methods with RESIT on a small network \textit{Asia} from the Bayesian network repository that has 8 nodes and 8 edges. Similarly, we simulated data sets with different percentages of nonlinear edges: $0\%, 20\%, 40\%, 60\%, 80\%, 100\%$. Four types of nonlinear relationships listed in Section~\ref{supp:nonlinear_functions} were simulated for empirical performance evaluation and comparison. Figure~\ref{fig:nonlinear_data_type} shows examples of different nonlinear patterns in the simulation. Each column corresponds to one type of nonlinear functions, and the two rows were simulated with randomly chosen parameters.

The regression method in RESIT can be selected from linear regression, generalized additive model (gam) and Gaussian process regression. Here we ran RESIT with gam and the default HSIC independence test for dependence measure.
We compared the two linear-NNCL algorithms and NNCL algorithm with RESIT. The $p$-value cutoff was set to $0.001$ for all three of our algorithms. 

\begin{figure}
    \centering
    \includegraphics[width=0.8\linewidth]{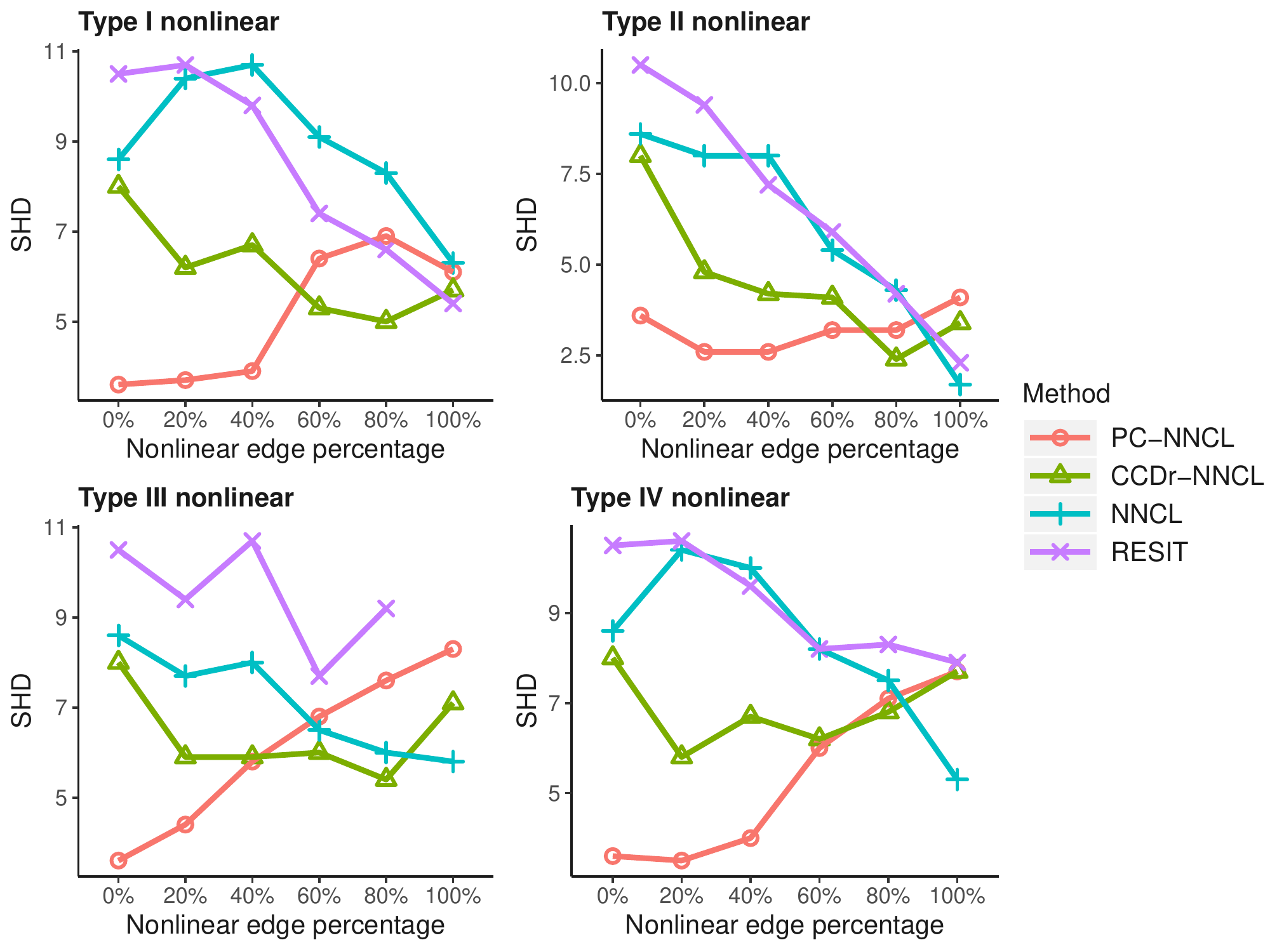}
    \caption{SHD comparison among four algorithms on different types of nonlinear models
    (Note: RESIT result missing for type \Romannumeral 3 with $100\%$ nonlinear edges due to an error in their code).}
    \label{fig:SHD_ANM_comparison}
\end{figure}

\begin{figure}
    \centering
    \includegraphics[width=0.8\linewidth]{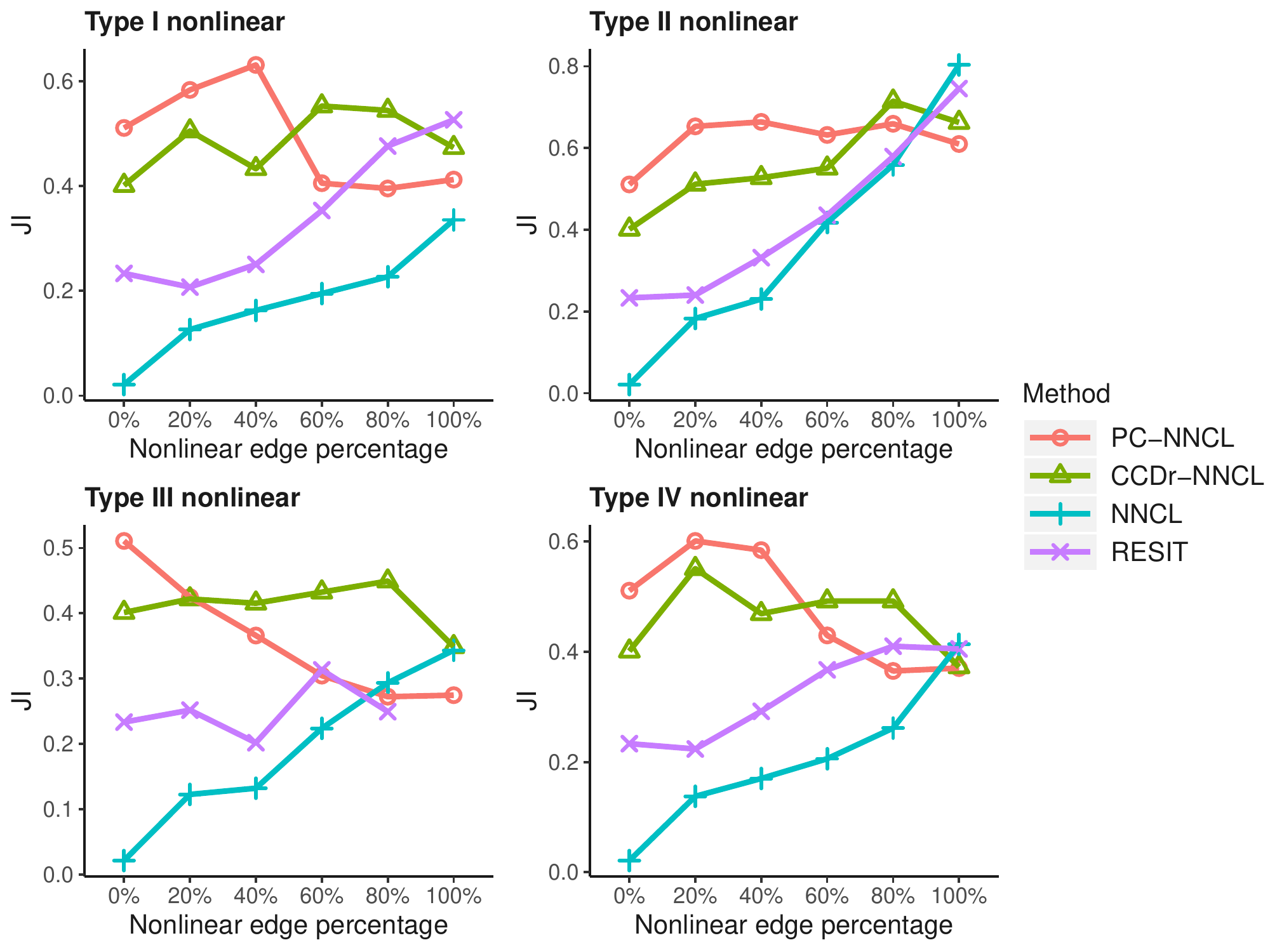}
    \caption{JI comparison among four algorithms on different types of nonlinear models (Note: RESIT result missing for type \Romannumeral 3 with $100\%$ nonlinear edges due to an error in their code).}
    \label{fig:JI_ANM_comparison}
\end{figure}

\begin{figure}[ht]
    \centering
    \includegraphics[width=0.7\linewidth]{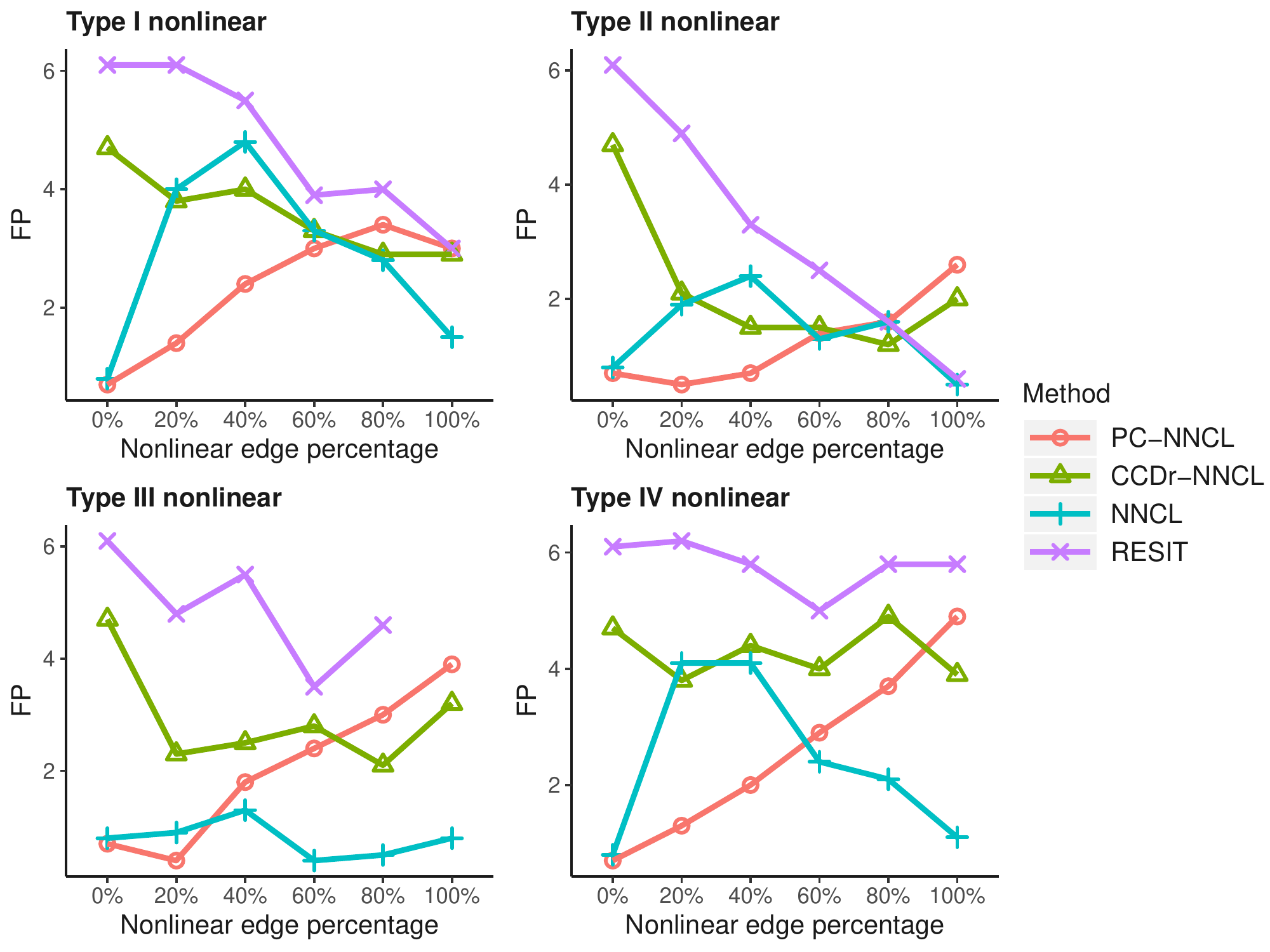}
    \caption{False positive comparison among four algorithms on different types of nonlinear models (note: RESIT result missing for type \Romannumeral 3 with $100\%$ nonlinear edges due to an error in their code)}
    \label{fig_supp:FP_ANM_comparison}
\end{figure}

Figures~\ref{fig:SHD_ANM_comparison} and \ref{fig:JI_ANM_comparison} show the performances of the four algorithms in terms of SHD and JI under different simulation settings. The ground truth we compared our results against was the true restricted CPDAG. Since RESIT always outputs a DAG, the performance of RESIT was compared to the true DAG instead.
True positive (TP) and false positive (FP) comparisons are provided in supplementary materials (Figure~\ref{fig_supp:TP_ANM_comparison}, \ref{fig_supp:FP_ANM_comparison}).

We observe from the plots that in general linear-NNCL algorithms performed similarly and showed the best results across all four cases of nonlinear relations. In particular, both PC-NNCL and CCDr-NNCL outperformed RESIT substantially for datasets with $\le 60\%$ of nonlinear edges and showed comparable accuracy with RESIT for cases with $\geq 80\%$ nonlinear edges. When there were fewer nonlinear edges, PC-NNCL algorithm (red lines) had lower SHDs and higher JI, which indicate better performance, while CCDr-NNCL algorithm (green lines) performed better in settings with a higher percentage of nonlinear edges. The NNCL algorithm (blue lines) showed lower accuracy when there were fewer nonlinear edges, but its performance improved greatly as the nonlinear percentage increased. The performance curves of RESIT (purple lines) exhibited similar trend as NNCL. RESIT was able to identify more correct edges than NNCL, however, at a cost of more superfluous edges (observed from the false positive curves) which led to a higher SHD between true and estimated graphs.

\begin{figure}
    \centering
    \begin{subfigure}[t]{0.4\linewidth}
    \centering\includegraphics[width=\linewidth]{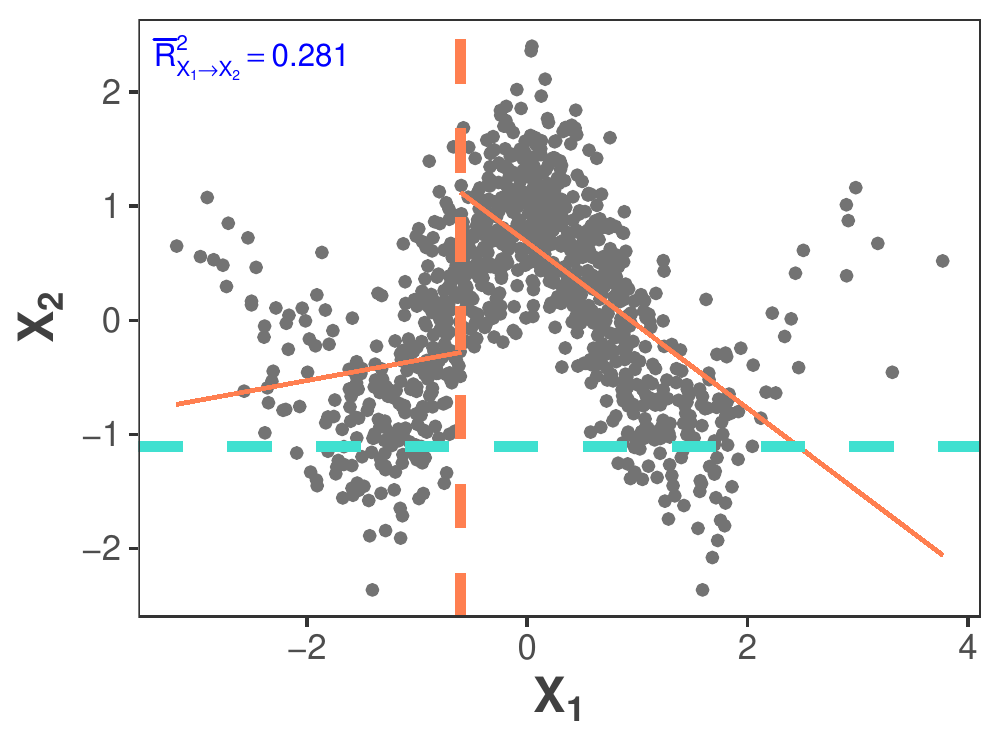} 
    \caption{\footnotesize Model fitting for $X_1\to X_2$ \label{fig:sufficiency_two_pieces_a}}
    \end{subfigure}
    \begin{subfigure}[t]{0.4\linewidth}
    \centering\includegraphics[width=\linewidth]{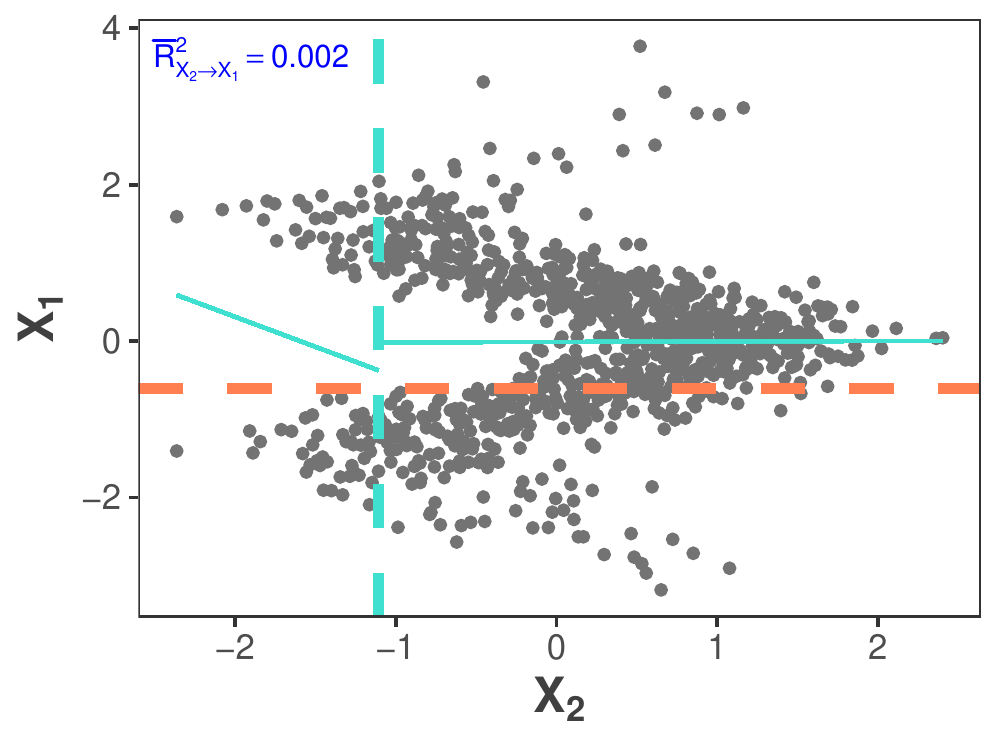}
    \caption{\footnotesize Model fitting for $X_2\to X_1$ \label{fig:sufficiency_two_pieces_b}}
    \end{subfigure}
    \caption{Example of complex non-invertible functions detected by our method with two-piece linear approximation.}
    \label{fig:sufficiency_two_pieces}
\end{figure}

The above results also confirm that the proposed NNCL algorithms were able to handle different types of nonlinear data. For instance, the nonlinear patterns in the second and fourth columns in Figure~\ref{fig:nonlinear_data_type} are obviously composed of multiple segments, yet our method had no problem detecting such non-invertible relationships using two-piece approximations as in Equation~\eqref{eq:piecewise_function}. Figure~\ref{fig:sufficiency_two_pieces} illustrates the detection of such complex non-invertible relationships by a two-piece linear model. 
The red dashed line in the figure is the estimated cut point of $X_1$ being the parent (i.e. $X_1\to X_2$), and the blue dashed line is the estimated cut point of $X_2$ being the parent (i.e. $X_2\to X_1$). The solid lines represent the fitted two-piece linear functions. We observe from Figure~\ref{fig:sufficiency_two_pieces_a} that the two pieces of functions captured the significant change in the nonlinear pattern. However, model fitting in the other direction $X_2 \to X_1$(Figure~\ref{fig:sufficiency_two_pieces_b}) failed to do so and resulted in a much smaller goodness of fit statistic $\bar{R}^2$. Therefore, a simple two-piece linear model allows us to identify more complex non-invertible relationships by capturing a single significant change in the pattern.
Of course, there are drawbacks using this simple procedure. Although we are able to successfully detect a non-invertible edge, the residuals obtained from the two pieces of linear models would be inaccurate and could affect the following detection if there are other undirected edges between $X_2$ and its neighbors. In such cases, a multiple-piece or more general nonlinear model fitting procedure is expected to be more powerful.

We also compared the computing time among the four algorithms in Table~\ref{tab:computing_time} on the data sets in this subsection. 
One sees that RESIT was quite time consuming for even a small network. The average computing time of RESIT was almost three times that of the other algorithms.

\begin{table}
    \caption{Comparison on computing time}    
        \centering
    \begin{tabular}{c|cccc}
    Method & PC-NNCL & CCDr-NNCL & NNCL & RESIT \\
    \hline
    Computing time in seconds & 10.059 & 11.944 & 12.699 & 35.043
    \end{tabular}
    \label{tab:computing_time}
\end{table}

\section{Application to ChIP-Seq data} \label{sec:application}

Although linear SEMs are commonly used in learning causal network structures, real-world data rarely satisfy a perfectly linear causal relationship. Therefore, assuming nonlinearity will help identify causal relationships from data. 
For example, some causal relations in biological data are expected to be nonlinear, exhibiting a piecewise trend. A gene $X$ may regulate gene $Y$ with a nonlinear functional relationship. When the expression level of $X$ is low, $X$ may have no effect on the expression of $Y$; but if the expression level $X$ passes certain threshold, it shows a strong positive regulation on $Y$. Similarly, the binding of transcription factors (TFs) to DNA may also show nonlinear causality. Transcription factors are a class of proteins that bind DNA in order to activate or suppress a downstream gene. The binding of one TF may stimulate the binding of another TF under a nonlinear dependence. Thus, it is an interesting and important problem to identify the causal network among the bindings of multiple TFs that work together in gene regulation.

In this section, we apply our methods to the ChIP-Seq data generated by \cite{ChenX2008}. 
The data set contains the DNA binding sites of 12 transcription factors in mouse embryonic stem cells: \textit{Smad1, Stat3, Sox2, Pou5f1, Nanog, Esrrb, Tcfcp2l1, Klf4, Zfx, E2f1, Myc,} and \textit{Mycn}. For each transcription factor, an association strength score, which is the weighted sum of the corresponding ChIP-Seq signal strength, was calculated for each of the 18,936 genes [\cite{Ouyang2009}]. Roughly speaking, this score can be understood as a measure of the binding strength of a TF to a gene. The genes with zero association scores were removed from our analysis. Accordingly, our observed data matrix, of size $n\times p= 8462\times12$, contains the association scores of 12 TFs over 8,462 genes. 
We aim to build a causal network that reveals how these $12$ TFs might affect each other's binding to genes.

Since there is no ground-truth for comparison, 
ten-fold cross validation was used to evaluate the performance of our methods. We first split the data into training and test sets, and ran a network learning method to obtain an estimated graph and associated parameters from training data. Then given an estimated network structure and the parameters, we calculated the likelihood of the test data set. Since the estimated graph was a PDAG, we extended the PDAG to an arbitrary DAG in the restricted equivalence class without creating any directed cycle or additional $v$-structures, and then used this DAG for estimating model parameters from training data and calculating test data likelihood. For simplicity, we postulated a quadratic function for each identified nonlinear edge, i.e. $f_{ki}(x)=a_k x + b_k x^2$ in Equation~\eqref{eq:lnSEM}, so that parameter estimation can be done by least-squares. The likelihood of test data was evaluated based on Gaussian error distributions.

From the simulation results in previous section, we find that CCDr-NNCL tends to have the best overall performance in different nonlinear settings. Therefore CCDr-based algorithms were chosen for this data analysis. The significance levels of the hypothesis tests and conditional independence tests in the NNCL steps were all set to $0.001$. Table~\ref{tab:cv} reports the results for CCDr and CCDr-NNCL averaging over 10 folds of cross validations. We see that the NNCL steps indeed identified on average 6.3 nonlinear edges and increased test data likelihoods compared to CCDr.

\begin{table}
\caption{Ten-fold cross validation results on ChIP-Seq data \label{tab:cv}}
    \centering
    \begin{tabular}{c|cccccc}
    Method & CCDr & CCDr-NNCL\\
    \hline
    average test data log-likelihood & -12081.1 & -11901.0\\
    average number of edges & 19.0 & 22.1\\
    average number of nonlinear edges & NA & 6.3 \\
    \hline
    \end{tabular}
    
\end{table}

\begin{figure}[ht]
    \centering
    \begin{subfigure}[t]{0.45\linewidth}
    \centering\includegraphics[width=\linewidth]{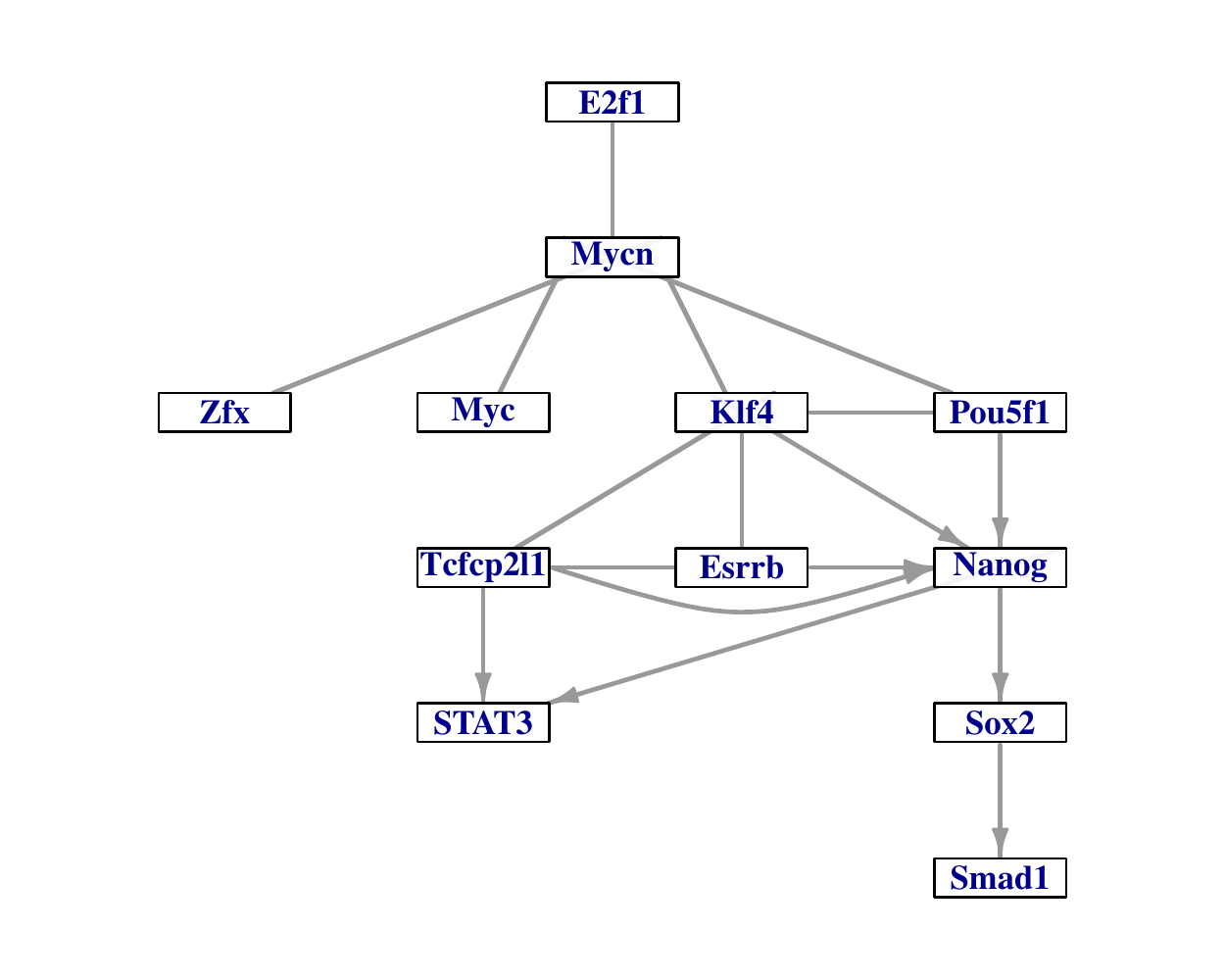}
    \caption{CCDr network (CPDAG)\label{fig:ChipSeq_a}}
    \end{subfigure}
    \begin{subfigure}[t]{0.45\linewidth}
    \centering\includegraphics[width=\linewidth]{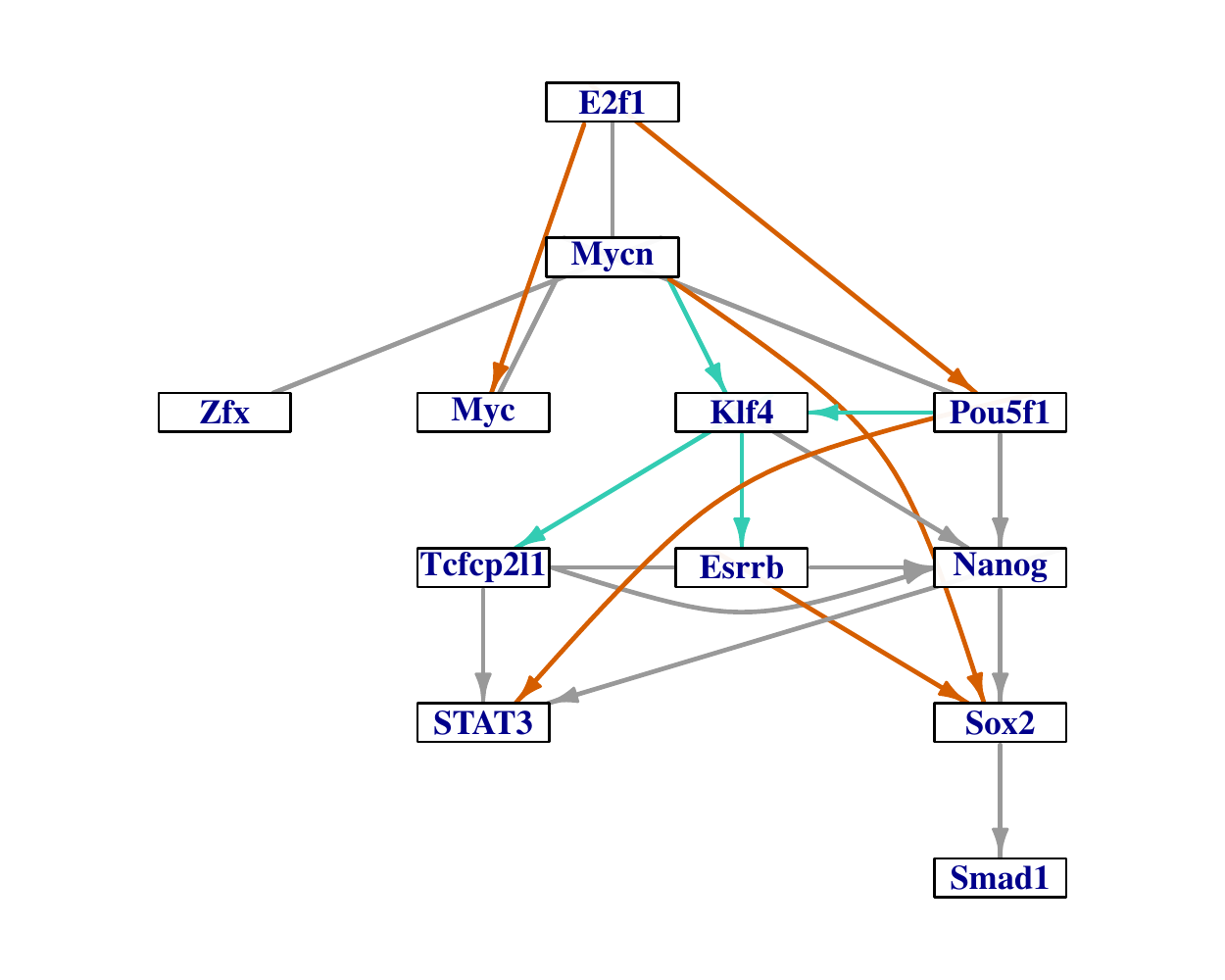}
    \caption{CCDr-NNCL network \label{fig:ChipSeq_b}}
    \end{subfigure}
    
    \begin{subfigure}[t]{0.45\linewidth}
    \centering\includegraphics[width=\linewidth]{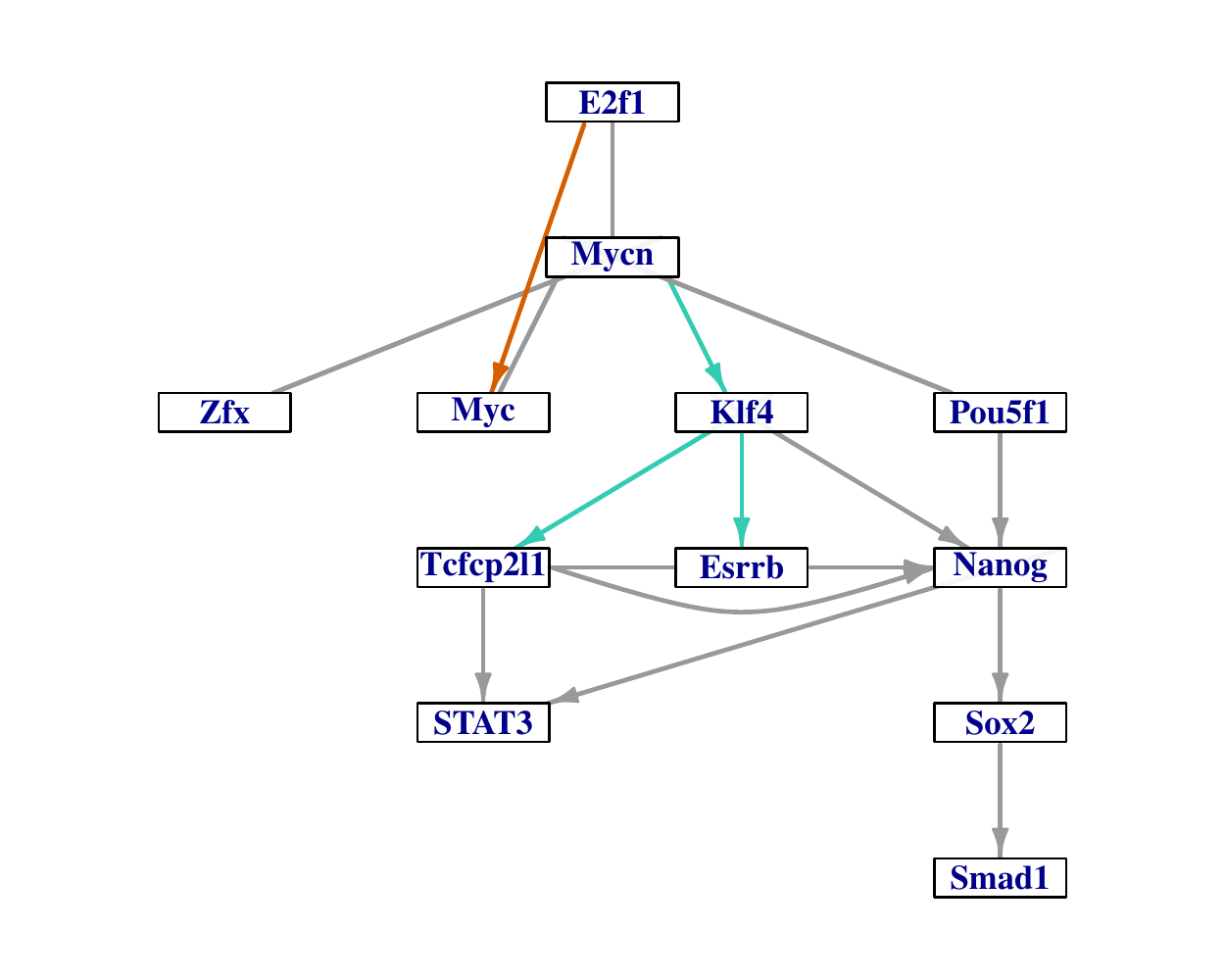}
    \caption{Consensus network \label{fig:ChipSeq_c}}
    \end{subfigure}
    \caption{TF binding causal networks estimated from ChIP-Seq data.}
    \label{fig:ChipSeq}
\end{figure}

The networks learned on the full dataset are shown in Figure~\ref{fig:ChipSeq}.
Figure~\ref{fig:ChipSeq_a} is the CPDAG of CCDr estimated network and Figure~\ref{fig:ChipSeq_b} is the network estimated by CCDr-NNCL. 
The green edges are nonlinear edges detected in the NNCL steps, and the red ones are nonlinear edges detected outside the skeleton using Algorithm~\ref{alg:outside_search}.
 
 To improve the stability of our estimated graph, a consensus network was constructed via bootstrap. Let ${\mathcal{G}}$ denote the estimated graph (PDAG) in Figure~\ref{fig:ChipSeq_b} by CCDr-NNCL and $\lambda$ be the tuning parameter used in the CCDr algorithm. First, we resampled the full dataset with replacement 100 times, and ran CCDr-NNCL with the same CCDr tuning parameter $\lambda$ on each bootstrap sample to obtain 100 estimated graphs $\{\mathcal{G}'_1, ..., \mathcal{G}'_{100}\}$. Second, we calculated a weighted adjacency matrix $W=(w_{ij})_{p\times p}$, where each entry of the adjacency matrix $w_{ij}$ recorded the percentage of the edge $i\to j$ appeared in the 100 estimated graphs $\{\mathcal{G}'_b, b=1,\ldots,100\}$. Finally, we constructed a consensus network using the weight matrix by the following rules. A directed edge $i\to j$ in ${\mathcal{G}}$ was kept if the weight $w_{ij} \geq 0.6$; a directed edge $i\to j$ in ${\mathcal{G}}$ was kept but changed to an undirected edge $i-j$ if $w_{ij}<0.6$ and $w_{ij} + w_{ji} \geq 0.6$; a directed edge $i\to j$ was deleted if neither of the above two conditions were satisfied. An undirected edge $i-j$ in $\mathcal{G}$ was kept if $w_{ij} + w_{ji} \geq 0.6$ and was deleted otherwise. 
The graph in Figure~\ref{fig:ChipSeq_c} is the consensus network so constructed with the same color code in Figure~\ref{fig:ChipSeq_b}.

It is well-known that two or more TFs may cooperate to regulate target genes. The work in \cite{Ouyang2009} suggests that \textit{E2f1, Myc, Mycn, Zfx} form one group of TFs (group I) that work together, and \textit{Pou5f1, Nanog, Sox2, Smad1, Stat3, Tcfcp2l1, Esrrb} form another group (II). We observe from the consensus network in Figure~\ref{fig:ChipSeq_c} that the group I TFs are more closely connected, and similarly group II TFs are also closely connected, consistent with their findings. \textit{Mycn} appears to be a point of junction in group I, and \textit{Nanog} seems to be an important connecting point in group II. Four nonlinear edges (green edges) were discovered from the CCDr skeleton and three of them were preserved in the consensus network. Only one edge out of the five edges detected outside the CCDr skeleton were preserved in the consensus network. Obviously the nonlinear edges detected within the skeleton were more stable. The results provide clues for nonlinear causal relations among TF binding events. Such pairs of TFs include \textit{E2f1 $\to$ Myc}, \textit{Mycn $\to$ Klf4}, \textit{Klf4 $\to$ Tcfcp2l1} and \textit{Klf4 $\to$ Esrrb}, in which \textit{Klf4} appears to interact with other TFs mostly in a nonlinear way. It would be interesting to further study the regulation roles of these TFs that showed nonlinear interactions.
Another observation from the consensus network is that \textit{Mycn} and \textit{Pou5f2} are the root causes of the binding of all group II TFs, while \textit{Stat3} and \textit{Smad1}, both in group II, are identified as sink nodes in all three estimated graphs.

 \section{Discussion} \label{sec:discussion}

Causal discovery from observational data is a crucial step to understanding causality in real world applications, especially when experiments are limited or infeasible. In this paper, we have demonstrated that non-invertible causal relationships can be identified from observational data. We started from the bivariate case, where the task was to decide the cause between two variables, and designed a test-based procedure to determine the causal direction. Furthermore, we extended the work to multivariate case and proposed an efficient algorithm which incorporates both linear structure learning and non-invertible SEMs to estimate the structure of a causal DAG.

We have tested and applied our methods on both simulated and real-world datasets. The simulation results indicate that by applying our NNCL algorithm, we can identify the causal directions of nonlinear edges with non-invertible relationships, and thus further reduce the Markov equivalence class estimated by traditional constraint-based or score-based DAG learning methods. Extensive numerical comparisons show that our linear-NNCL algorithms are able to handle different types of nonlinear relationships and outperform the RESIT algorithm in most cases. The application to ChIP-Seq data highlights the utility of incorporating nonlinear SEMs in learning causal networks.

Several topics will be studied in future work.
 The two-piece linear model will lead to a loss of accuracy when fitting more complex nonlinear causal relations. For example, using residuals from the two-piece model may result in false negatives in non-invertibility tests. Generalization of our model from two pieces to multiple pieces can help improve model fitting and edge detection of our algorithms for more complicated data.
 The hypothesis test for causal direction we proposed in this paper is based on sample correlation coefficients. 
 Other possible statistics await to be explored in the future.
Finally, more theoretical work can be developed to study the large-sample properties of our methods.

\begin{supplement}
 \sname{S1}
 \stitle{Conditional independence test}
 \slink[doi]{}
 \slink[url]{url}
 \sdescription{Details of Conditional independence test in Algorithm~\ref{alg:NNCL}}
\end{supplement}
\begin{supplement}
 \sname{S2}
 \stitle{Nonlinear functions}
 \slink[doi]{}
 \slink[url]{url}
 \sdescription{Nonlinear functions used in Section~\ref{ss:ANM_comparison}}
\end{supplement}
\begin{supplement}
 \sname{S3}
 \stitle{Supplementary figures}
 \slink[doi]{}
 \slink[url]{url}
 \sdescription{
 Additional figures of comparisons in simulation study
 }
\end{supplement}



\bibliographystyle{imsart-nameyear} 
\bibliography{references}       


\clearpage

\beginsupplement

\begin{center}
\textbf{Supplementary Material}
\end{center}

\section{Conditional independence test in Algorithm~\ref{alg:NNCL}}
\label{supp:cond_ind_test}

There are many different methods to test the conditional independence of two variables $(X, Y)$ given a set of conditioning variables $\textbf{Z}$. Here in step \ref{alg_nncl_ln:conditions}(ii) of the algorithm, we use the partial correlation $\rho_{[XY|\textbf{Z}]}$ to test the conditional independence of $(X, Y) | \textbf{Z}$. The partial correlation between $X$ and $Y$ is the correlation between the residuals $e_X$ and $e_Y$ resulting from the linear regression of $X$ on $\textbf{Z}$ and $Y$ on $\textbf{Z}$, Fisher's z-transform of the partial correlation can used to test if the sample partial correlation $r_{[XY|\textbf{Z}]}$ implies a true population partial correlation of 0.
$$z(r_{[XY|\textbf{Z}]}) = \frac{1}{2} \log\frac{1+r_{[XY|\textbf{Z}]}}{1-r_{[XY|\textbf{Z}]}} \sim N\bigg(0, \frac{1}{\sqrt{N - |\textbf{Z}| -3}}\bigg)$$
The null hypothesis $\rho_{[XY|\textbf{Z}]} = 0$ is rejected at significance level $\alpha$ if 
$$\sqrt{N-|\textbf{Z}|-3} | \cdot z(r_{[XY|\textbf{Z}]})| > \Phi^{-1}(1-\alpha/2).$$

\section{Nonlinear functions used in Section~\ref{ss:ANM_comparison}}

\label{supp:nonlinear_functions}

\begin{enumerate}
\item Type \Romannumeral 1: $y = \pm ax^2 \pm bx + e$; 
\item Type \Romannumeral 2: $y = \pm\cos(ax) + e$;
\item Type \Romannumeral 3: $y = \pm ax^3 \pm bx^2 + e$;
\item Type \Romannumeral 4: $y = \pm\tanh{x} \pm \cos{ax} \pm x^2 + e$.
\end{enumerate}

In our experiments, $x, e$ were simulated from $N(0,1)$ and $a, b$ were drawn randomly from Unif(0.3, 4).


\newpage

\begin{figure}[ht]
    \centering
    \includegraphics[width=0.7\linewidth]{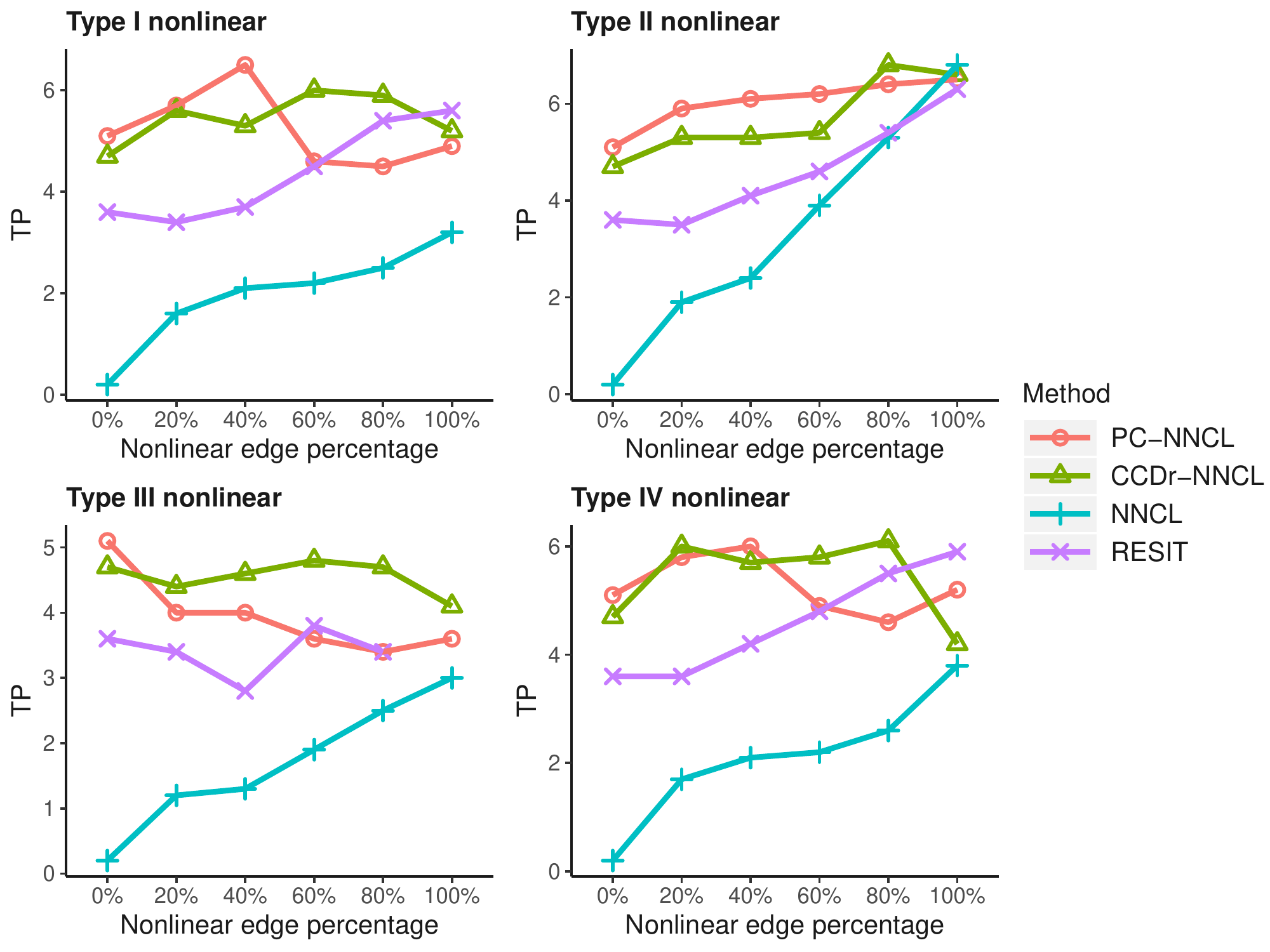}
    \caption{True positive comparison among four algorithms on different types of nonlinear models (note: RESIT result missing for type \Romannumeral 3 with $100\%$ nonlinear edges due to an error in their code)}
    \label{fig_supp:TP_ANM_comparison}
\end{figure}

%
%

%
%
\end{document}